\documentclass[10pt,twocolumn,letterpaper]{article}

\usepackage{titling}
\usepackage{cvpr}
\usepackage{times}
\usepackage{epsfig}
\usepackage{graphicx}
\usepackage{amsmath}
\usepackage{amssymb}
\usepackage[keeplastbox]{flushend}

\usepackage[format=plain,labelformat=simple,labelsep=period,font=small,skip=4pt,compatibility=false]{caption}
\usepackage[font=scriptsize,skip=2pt]{subcaption}
\usepackage{tabularx}
\usepackage{booktabs}
\usepackage{etoolbox,siunitx}
\robustify\bfseries
\DeclareSIUnit[number-unit-product={}]{\percent}{\%}
\usepackage[super]{nth}
\usepackage{newtxtt}
\usepackage{setspace}
\usepackage{fancyhdr}

\usepackage[breaklinks=true,bookmarks=false]{hyperref}

\usepackage{multibib}
\newcites{supp}{References}

\usepackage[capitalize]{cleveref}
\crefname{section}{Sec.}{Secs.}

\cvprfinalcopy 
\def\cvprPaperID{5138} 

\clearpage{}
\usepackage{xspace}
\newcommand{\Eq}{Eq.\@\xspace}

\newcommand{\R}{\ensuremath{\mathbb{R}}}

\newcommand{\mv}[1]{\ensuremath{\mathbf{#1}}}

\newcommand{\f}{\mathbf{f}}
\newcommand{\p}{\mathbf{p}}
\newcommand{\cc}{\mathbf{c}}
\newcommand{\vv}{\mathbf{v}}
\newcommand{\bb}{\mathbf{b}}

\newcommand{\W}{\mathbf{W}}

\newcommand{\myparagraph}[1]{\smallskip\noindent\textbf{#1}\hspace{0.5em}}

\usepackage{pifont}
\newcommand{\cmark}{\ding{51}}
\newcommand{\xmark}{\ding{55}}

\newcommand{\transp}{\text{T}}

\clearpage{}

\ifcvprfinal\fi
\begin{document}

\title{Probabilistic Pixel-Adaptive Refinement Networks}

\author{Anne S. Wannenwetsch$^{1,2}$\thanks{This work was done at TU Darmstadt prior to Anne S. Wannenwetsch joining Amazon.} \qquad Stefan Roth$^2$ \\
$^1$Amazon, Germany \qquad $^2$ TU Darmstadt, Germany}

\maketitle
\thispagestyle{fancy}

\begin{abstract}
   
Encoder-decoder networks have found widespread use in various dense prediction tasks.
However, the strong reduction of spatial resolution in the encoder leads to a loss of location information as well as boundary artifacts.
To address this, image-adaptive post-processing methods have shown beneficial by leveraging the high-resolution input image(s) as guidance data.
We extend such approaches by considering an important orthogonal source of information: the network's confidence in its own predictions.
We introduce \emph{probabilistic pixel-adaptive convolutions (PPACs)}, which not only depend on image guidance data for filtering, but also respect the reliability of per-pixel predictions.
As such, PPACs allow for image-adaptive smoothing and simultaneously propagating pixels of high confidence into less reliable regions, while respecting object boundaries.
We demonstrate their utility in refinement networks for optical flow and semantic segmentation, where PPACs lead to a clear reduction in boundary artifacts.
Moreover, our proposed refinement step is able to substantially improve the accuracy on various widely used benchmarks.
 \end{abstract}

\section{Introduction}
\label{sec:intro}
Convolutional neural networks (CNNs) have become a standard tool in computer vision.
Especially in dense prediction tasks \cite{Chen:2018:EAS, Long:2015:FCN, Ronneberger:2015:UCN, Yin:2019:HDD},
encoder-decoder or pyramid-structured CNNs are a common choice.
While originating in unsupervised learning \cite{Hinton:1993:AMD}, such architectures have become popular also in supervised settings.
The encoder builds a powerful feature representation, reducing the spatial resolution of the inputs to aggregate global information \cite{Long:2015:FCN}.
The decoder takes the feature representation from the bottleneck, enlarges its size, and transforms it into the desired output, \eg a segmentation map or optical flow field.

\begin{figure}[t]
	\centering
	\includegraphics[width=\linewidth]{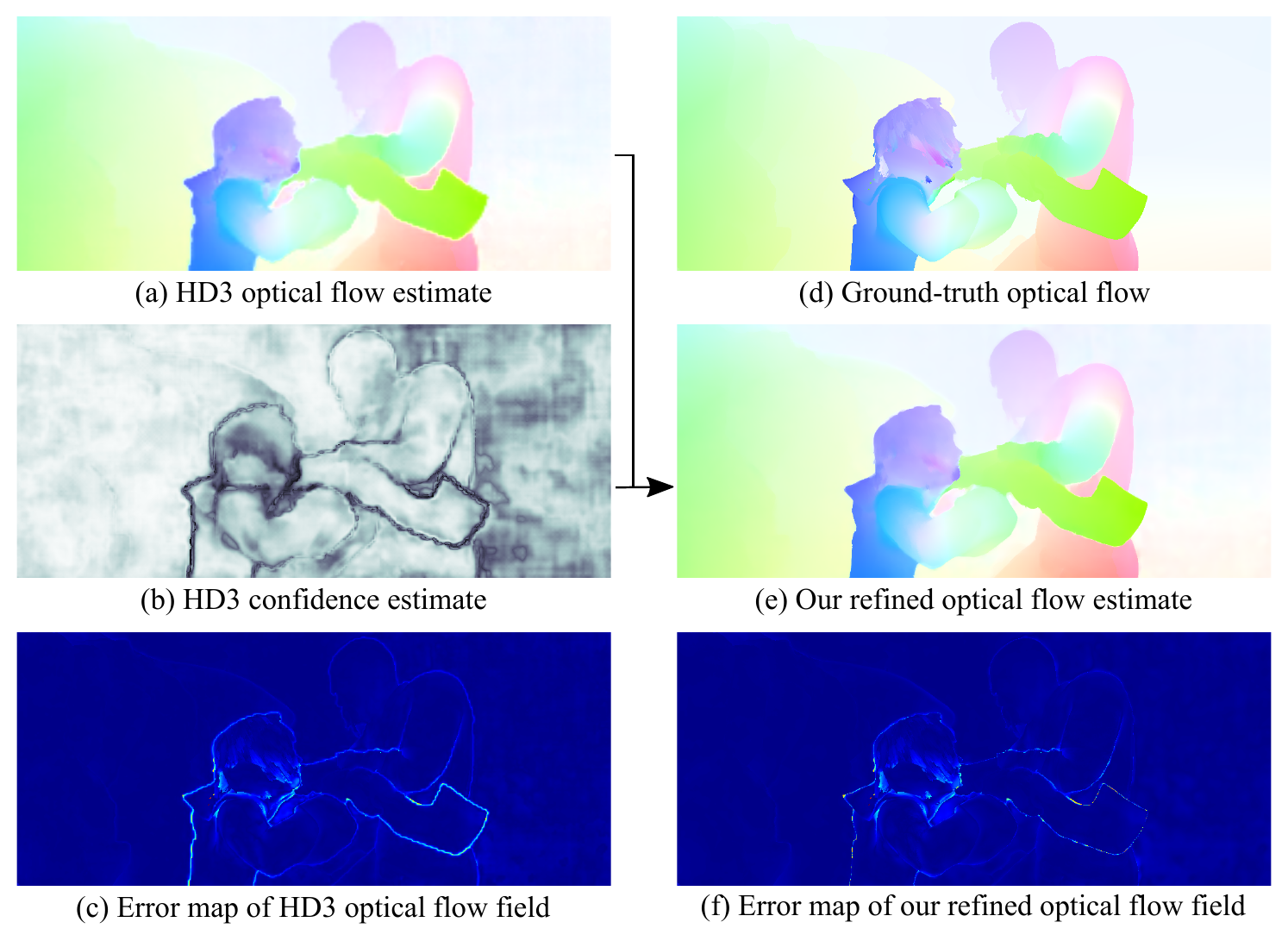}
	\caption{Our PPAC refinement method leverages the close relationship between estimated confidences and prediction errors to refine and improve the prediction itself, here the optical flow field.}
\label{fig:figure1}
\vspace{-0.5em}
\end{figure}

While downsampling in the encoder increases the receptive field and allows to deal with large image sizes, it also leads to a drastic loss in spatial resolution.
As such, valuable location information is lost and boundary artifacts can arise \cite{Harley:2017:SCN, Long:2015:FCN}, \eg segmentation maps that are misaligned \wrt the underlying objects.
Moreover, the decoder typically yields low-resolution outputs and simple components are used to upscale predictions to the input size.
This often results in blurry outputs since estimates of different objects are combined, \eg motion from background and foreground objects is mixed as can be seen in \cref{fig:figure1}.

Several approaches have been proposed to reduce these disadvantages, \eg skip connections \cite{Long:2015:FCN,Ronneberger:2015:UCN} or densely connected blocks \cite{Huang:2017:DCC, Jegou:2017:OHL}.
Additionally, different types of generalized convolutions, taking into account a high-resolution RGB guidance image,
have shown beneficial as part of the decoder \cite{Hui:2018:LLC, Hur:2019:IRR2}
or in a separate upsampling and/or refinement step \cite{Jampani:2016:LSH, Pan:2019:SVL, Wannenwetsch:2019:LTS, Wu:2018:FET2}.
Many of the above approaches require a large number of additional parameters, are computationally expensive, or restrictive \wrt to the filtering method and the applicable guidance data.
Most recently, pixel-adaptive convolutions (PACs) were introduced by  Su \etal \cite{Su:2019:PAC}.
PACs combine spatially-invariant convolution weights with a content-adaptive kernel that depends on guidance data.
In \cite{Su:2019:PAC}, PACs are shown to yield state-of-the-art results in joint upsampling tasks.

In this paper, we argue that we can leverage another source of information
for refinement that is complementary to the input image: \emph{the uncertainty of each pixel's estimate}.
For dense classification networks, this quantity is generally provided implicitly.
For instance, segmentation networks usually output log-probabilities of the classes, which are passed through an $\arg\max$ operation at test time.
Even though these uncertainties might not be well calibrated \cite{Guo:2017:OCM}, we argue that they still contain valuable information for refining the predictions.
Beyond classification, explicit uncertainty estimates are also gaining increased attention for dense regression problems such as geometry \cite{Novotny:2018:SSL} or motion estimation \cite{Gast:2018:LPD,Ilg:2018:UEM,Yin:2019:HDD}.
They are, for example, helpful for applications in which the reliability of network estimates is crucial, \eg in autonomous driving.
Here, we show that we can also leverage them to refine the regression output itself.

\cref{fig:figure1} shows an optical flow field estimated by the probabilistic HD3 method \cite{Yin:2019:HDD}, as well as the corresponding confidence map and endpoint error per pixel.
We observe that regions of high uncertainty (\emph{b}, dark gray) correspond quite well with large errors (\emph{c}, dark red).
When applying post-processing to the network output, it seems desirable to take the available pixel uncertainty into account.
As such, only reliable pixels should be spatially propagated while uncertain pixels can be replaced.
To allow for probability-aware\footnote{By \emph{probability} we refer to a measure that approximates or summarizes the marginal posterior over the network's estimates, \eg by taking the marginal posterior of the chosen prediction value. To ease readability, the terms probability, confidence, (un)certainty, and reliability will be used interchangeably in the paper.} filtering, we propose \emph{probabilistic pixel-adaptive convolutions (PPACs),} and therefore extend the adaptive convolution operation of \cite{Su:2019:PAC}.
The kernels of PPACs and thus the filtering output vary dependent on two properties:
guidance data, \eg the input image, as well as a probability map estimated by the deep network, either inherently or explicitly.

This paper focuses on the application of PPACs for the refinement of outputs from dense prediction networks,
illustrated with the tasks of optical flow estimation and semantic segmentation.
Therefore, we introduce a \emph{PPAC refinement network}, which leverages RGB guidance data and probabilities for several content- and probability-adaptive convolutions.
For both tasks, PPACs not only allow to improve estimates at boundaries but also remove outliers of low reliability.
As shown in \cref{fig:figure1}, blurry edges in the flow field \emph{(a)} are transformed into crisp boundaries and the overall prediction is smoothed \emph{(e)}.
Along with the visual improvements, PPAC refinement leads to a clear accuracy gain in optical flow and semantic segmentation.
For instance, PPACs substantially improve state-of-the-art HD3 \cite{Yin:2019:HDD} optical flow estimates on the widely used KITTI 2012 and 2015 datasets.
Our proposed PPAC-HD3 method ranks \nth{1} among published optical flow approaches\footnote{All rankings at the time of publication.} on both benchmarks, improving the outlier rate by $\sim$11.1$\%$ and $\sim$7.5$\%$ over the underlying baseline method. 
\section{Related Work}
\paragraph{Probabilistic deep networks.}
Combining probabilistic approaches with deep networks is an active field of research, which is pursued to cope with model and/or input uncertainty \cite{Kendall:2017:WUD}.
As such, we can only provide a rough summary and refer to the cited references for a broader overview.

Bayesian neural networks \cite{Blundell:2015:WUN, Graves:2011:PVI, Hernandez:2015:PBS, Kingma:2015:VDL, Louizos:2017:MNF, Wu:2019:DVI} often learn parametric distributions over the network weights to capture model uncertainty.
The predictive distribution over the outputs is obtained by taking an expectation over the weights through approximate inference \cite{Blundell:2015:WUN, Hernandez:2015:PBS}.
However, Bayesian neural networks introduce many additional parameters and are not always easy to handle in practice \cite{Lakshminarayanan:2017:SSP}.

Sampling-based approaches, \eg \cite{Ayhan:2018:TDA, Gal:2016:DBA, Kendall:2017:WUD}, are often simpler to apply and include a random component, such as dropout, in the network structure.
At test time, the predictive uncertainty is computed as Monte Carlo estimates from several network passes.
Similarly, an ensemble of networks can be trained and combined at test time \cite{Huang:2017:SET, Lakshminarayanan:2017:SSP}.
A major drawback of both avenues is the increased runtime as they require multiple forward passes.

Another line of research uses deep networks to output the parameters of an assumed predictive distribution, either directly \cite{Kendall:2017:WUD, Nix:1994:EMV} or by propagation of input uncertainty \cite{Gast:2018:LPD, Wang:2016:NPN}.
There, the difficulty is to find a parametric distribution that is sufficiently easy to handle in practice and appropriately describes the quantity of interest.

Beyond such general purpose probabilistic treatments, probabilistic networks have also been developed in the context of specific vision problems.
Yin \etal~\cite{Yin:2019:HDD} propose a method to aggregate correspondence uncertainty in the context of optical flow and stereo matching through various spatial scales.
In \cite{Ilg:2018:UEM}, a multi-hypothesis network for optical flow estimation is trained to output an ensemble at once.
Novotny \etal~\cite{Novotny:2018:SSL} use uncertainty estimates obtained with probabilistic losses to predict the reliability of descriptors.

\myparagraph{Content-adaptive convolutions.}
One category of content-adaptive convolutions adjusts the sampling location of neighbor pixels \cite{Dai:2017:DCN, Jeon:2017:ACL}.
Deformable convolutions \cite{Dai:2017:DCN} predict data-dependent offsets to determine at which locations neighboring pixels should be sampled for a spatially-invariant convolution.
Another line of research adjusts the convolution weights \cite{Jia:2016:DFN, Wu:2018:DFL} of standard convolutions.
Dynamic filter networks \cite{Jia:2016:DFN} use a subnetwork to predict location-specific weight kernels, which have already shown benefits for optical flow estimation, \eg \cite{Hui:2018:LLC, Hur:2019:IRR2}.
A common drawback is the significant amount of additional parameters, which increases the risk of overfitting -- especially if only a limited amount of training data is available.

Several works, therefore, approach content-adaptive convolutions in a more constrained setting.
Spatial transformers \cite{Jaderberg:2015:STN} as well as CARAFE \cite{Wang:2019:CCA} rearrange features in a content-adaptive way with a global or local transformation before performing the convolution itself.
However, they remain restricted to a 2D grid structure.
\cite{Jampani:2016:LSH, Wannenwetsch:2019:LTS} perform image-adaptive convolutions with predefined or learned features in the high-dimensional permutohedral lattice \cite{Adams:2010:FHF}.
Such convolutions are computationally expensive and thus not suitable for fast processing.
\cite{Harley:2017:SCN} incorporates semantics by learning input-dependent attention masks but requires object classes for \mbox{(pre-)}training.
Deep guided filters \cite{Wu:2018:FET2} extend the classical guided filter \cite{He:2010:GIF} to learned guidance data.
In \cite{Pan:2019:SVL}, the parameters for spatially-variant linear representation models of the guided filter are learned with a CNN.
In both approaches only 1D guidance data can be used for each output channel.
Deep joint image filtering \cite{Li:2019:JIF} applies standard convolutions to the concatenation of pre-processed guidance data and estimates, but misses explicit knowledge on the relation of both components.

We base our approach on pixel-adaptive convolutions (PACs) \cite{Su:2019:PAC}.
Here, the convolution kernels are split into a fixed weight as well as a location-specific component that depends upon a feature embedding learned from guidance data.
PACs have a small computational overhead and are easy to train in practice.
Nevertheless, like most adaptive filtering approaches, PACs do not allow to explicitly leverage knowledge about the reliability of filter inputs.
While such information can be included as guidance data for the content-adaptive part of the weight kernel, our explicit probabilistic formulation leads to a significant performance gain.

\myparagraph{Probabilistic joint filtering.}
There are only few filtering approaches that jointly consider guidance data as well as probabilities.
Different filtering methods have been applied to refine semantic segmentations \cite{Sakaridis:2019:GCM}, optical flow \cite{Mai:2017:OFR, Vogel:2018:LEB}, and especially depth
\cite{Ham:2018:RGI, Knoebelreiter:2019:LCS, Mueller:2010:ACT}.
However, these approaches are task-specific, tailored to certain filtering methods, and/or rely on time-intensive iterative approaches.
\cite{Gidaris:2017:DRR} replaces unreliable pixels using a network that takes uncertainties and guidance data as input, but uncertainties are not explicitly leveraged to improve estimates.
The Fast Bilateral Solver \cite{Barron:2016:FBS} and the Domain Transform Solver \cite{Bapat:2019:DTS} allow to perform fast, edge-aware optimization on different estimates.
They leverage uncertainty by requiring a closer connection between inputs and outputs for more reliable pixels.
As these approaches optimize a predefined objective, their flexibility is restricted.
Moreover, \cite{Bapat:2019:DTS, Barron:2016:FBS} cannot backpropagate into the features used to determine the pixel similarity.
Closest to ours with regard to probabilistic joint filtering are \cite{Hoerentrup:2014:CAG, Jachalsky:2010:CER}, where confidences are used to extend the bilateral and the guided filter, respectively, by weighing a pixel's importance with its confidence.
However, both methods rely on predefined features for pixel similarity, hand-crafted reliability measures, and fixed filter kernels.

In comparison to previous work, our approach is very general as pixel feature embeddings, the filter weights, as well as a pre-processing of the probabilities are learned from data.
Moreover, the proposed approach is fast and easily integrable into different task-specific neural networks. 
\section{Probabilistic Pixel-Adaptive Convolutions}
We begin by presenting pixel-adaptive convolutions (PACs) as introduced in \cite{Su:2019:PAC}.
We then propose an advanced normalization approach for pixel-adaptive convolutions and finally extend PACs to allow for probabilistic filtering.

\subsection{Pixel-adaptive convolutions}
Assume, we aim to perform a convolution with neighborhood size $s$ that transforms features $\vv \in \R^d$ into features $\tilde{\vv} \in \R^{d'}$.
We denote the corresponding convolution weights as tensor $\W \in \R^{d' \times d \times s \times s}$ and the bias term as $\bb \in \R^{d'}$.
Following the notation of \cite{Su:2019:PAC}, the output of a standard convolution at pixel $i$ is then given as
\begin{equation}
\tilde{\vv}_i = \sum_{j \in \mathcal{N}(i)} \W \left[ \p_i - \p_j \right] \vv_j + \bb,
\end{equation}
where $\mathcal{N}(i)$ denotes the $s \times s$ neighborhood of the pixel.
The vectors $\p_i$ and $\p_j$ represent the 2D pixel positions.
$\W \left[ \p_i - \p_j \right] \in \R^{d' \times d}$ corresponds to the 2D slice from weight tensor $\W$, evaluated at position $\p_i - \p_j$.

PACs generalize such spatially-invariant convolutions by augmenting the convolution weight with an additional location-adaptive component $K\big(\f_i, \f_j \big)$.
In \cite{Su:2019:PAC}, the vectors $\f_i$ and $\f_j$ are denoted as \emph{pixel features} and characterize the pixels $i$ and $j$, respectively.
For instance, one could use the RGB components of a guidance image as feature $\f_{(\cdot)}$, as is done in the bilateral filter \cite{Tomasi:1998:BFF}.
However, more advanced features learned from data have shown to be advantageous \cite{Su:2019:PAC}.
The function $K(\f_i, \f_j) = K(\f_i - \f_j)$ is a (fixed) kernel, which evaluates the difference between $\f_i$ and $\f_j$.
If pixels $i$ and $j$ show similar characteristics, $K(\f_i, \f_j)$ weighs the corresponding values $\vv_j$ more than the ones of a more deviating pixel.
Various choices for $K(\cdot, \cdot)$ are possible; we will apply a Gaussian RBF kernel in the following, \ie
\begin{equation}
K\big(\f_i, \f_j \big) = e^{- \frac{1}{2} (\f_i - \f_j)^{\transp} (\f_i - \f_j)}.
\label{eq:kernel_func}
\end{equation}
The PAC convolution \cite{Su:2019:PAC} is then defined as 
\begin{equation}
\tilde{\vv}_i = \sum_{j \in \mathcal{N}(i)} K \big(\f_i, \f_j \big) \cdot\W \left[ \p_i - \p_j \right]  \vv_j + \bb.
\label{eq:pac}
\end{equation}
The same feature kernel $K$ is used for all input channels, while the weight $\W$ differs dependent on the spatial location within the mask and for each feature channel.

\subsection{Advanced normalization step}
\label{sec:advanced_normalization}

PACs are a powerful tool for deep dense prediction architectures, but they are not without challenges.
One major issues is the fact that the number of closely related pixels, \ie pixels $j$ that show a high value of $K(\f_i, \f_j)$, varies across the image.
This is natural and even desirable, since only a restricted neighborhood region should be taken into account at object boundaries or similar.
Nevertheless, applying PACs in different neighborhoods should lead to results with the same output magnitude as long as the input variables are equal.
Otherwise, learning of convolution weights might be difficult since the output values are inevitably smaller at boundaries or in highly structured areas.

\begin{figure}[t]
	\centering
	\includegraphics[width=\linewidth]{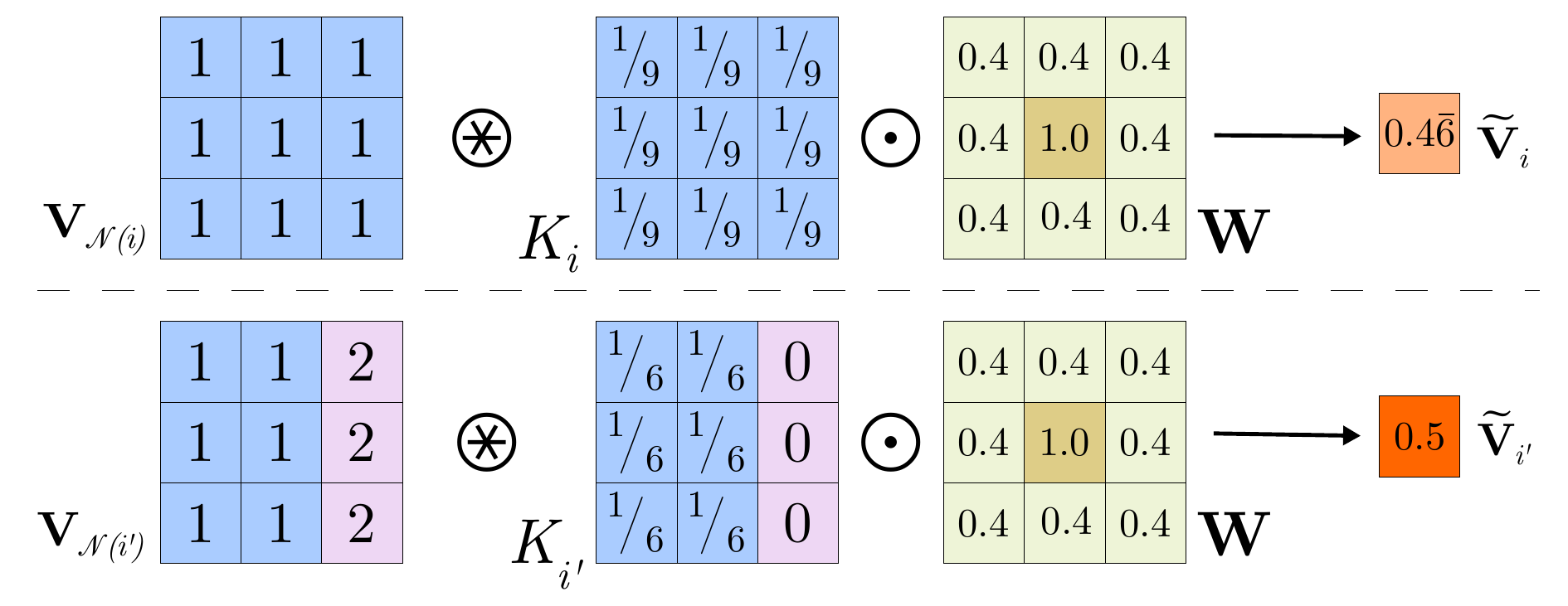}
	\caption{PAC normalization \wrt kernel only (see text).}
\label{fig:normalization_kernel_weights}
\end{figure}

\myparagraph{Basic scheme.}
The implementation of PACs provides an option to normalize kernels such that $\sum_{j} K(\f_i, \f_j) = 1$ for all pixels $i$.
However, we argue that such a \emph{kernel normalization} is not sufficient.
Consider the illustration of two pixels $i$ and $i'$ and their neighborhoods $\mathcal{N}(i)$ and $\mathcal{N}(i')$ in \cref{fig:normalization_kernel_weights}.
For simplicity, we assume equal input values for pixels of the same object.
Pixel $i$ is part of a homogeneous area and the kernel function leads to an equal distribution of the kernel weights.
In contrast, the neighborhood of pixel $i'$ contains also elements from a different object.
Here, the kernel weights are only distributed over the elements from the same object as pixel $i'$.
Even though both kernels sum to one, their convolution with an exemplary weight $\W$ leads to clearly different results.
As mentioned above, this complicates the learning of PACs.

\myparagraph{Our advanced scheme.}
We address this issue with an advanced normalization scheme.
To that end, we adapt the normalization from \cite{Wannenwetsch:2019:LTS} to the context of PACs.
An auxiliary array $\vv^{aux}\!=\!\mv{1}$ with constant value $1$ is defined and passed through the filtering step in \cref{eq:pac}:
\begin{equation}
\tilde{\vv}^{aux}_i = \sum_{j \in \mathcal{N}(i)} K \big(\f_i, \f_j \big) \cdot\W \left[ \p_i - \p_j \right]  \cdot \mathbf{1}.
\label{eq:normalization_pass}
\end{equation}
In slight abuse of notation, we now denote by $\tilde{\vv}_i$ the PAC output in \cref{eq:pac} \emph{before} adding the bias term.
Then, the normalized convolution output $\tilde{\vv}_{i,\text{norm}}$ is given by
\begin{equation}
\tilde{\vv}_{i,\text{norm}}=\tilde{\vv}_i / \tilde{\vv}^{aux}_i + \bb.
\label{eq:normalized_pac}
\end{equation}
With the novel normalization, not only the number of similar pixels is taken into account but also their weighting by $\W$.
However, the normalization becomes invalid as soon as the weight $\W$ becomes negative \cite{Wannenwetsch:2019:LTS}.
We thus follow \cite{Wannenwetsch:2019:LTS} to introduce an additional \emph{normalization weight} $\W'$, which replaces $\W$ in the normalization convolution in \cref{eq:normalization_pass}.
$\W'$ is ensured to remain positive and set to the same initialization as $\W$.
During training, we independently update $\W$ as well as $\W'$ using regular gradient-based optimization and thus omit to explicitly enforce their similarity.

Reconsidering the example in \cref{fig:normalization_kernel_weights}, both output values remain close when using our advanced normalization with the appropriate $\W'\approx\W$.
We will show in \cref{sec:exp_optical_flow} that the proposed scheme leads to superior results in practice.

\subsection{Probabilistic pixel-adaptive convolutions}
While the definition of PACs allows for more advanced filtering than a standard convolution, the approach is still restricted.
The kernel function $K(\cdot,\cdot)$ in \cref{eq:kernel_func} only takes differences of pixel features as input, thus excluding properties that cannot be reasonably expressed as such.
Consider, for instance, that we have a per-pixel probability alongside the estimate.
In this case, it seems beneficial to consider such information during filtering to improve unreliable estimates.
As $K(\cdot,\cdot)$ rewards similar pixel features, neighbors with a similar level of reliability are more closely connected.
However, this seems counterintuitive, given that \emph{the values of reliable pixels should particularly propagate to neighbors with a very different, \ie low, confidence}.

Therefore, we extend PACs such that we are able to perform convolutions that also take unary properties -- especially probabilities -- into account.
Let $c_j$ describe the confidence assigned to a certain pixel location $j$.
For consistency, we assume that only one confidence estimate is given per spatial location.\footnote{An extension to individual confidences per channel is straightforward.}
Similar to \cite{Jachalsky:2010:CER}, we then propose to define a \emph{probabilistic pixel-adaptive convolution (PPAC)} as:
\begin{equation}
\tilde{\vv}_i = \sum_{j \in \mathcal{N}(i)} c_j \cdot K \big(\f_i, \f_j \big) \cdot \W \left[ \p_i - \p_j \right]  \vv_j + \bb.
\label{eq:ppac}
\end{equation}
Here, each pixel value is not only weighted by its distance to the center pixel but also with its individual confidence.

To illustrate our proposed approach, consider \cref{fig:probabilistic_pac}:
\begin{figure}[t]
	\centering
	\includegraphics[width=\linewidth]{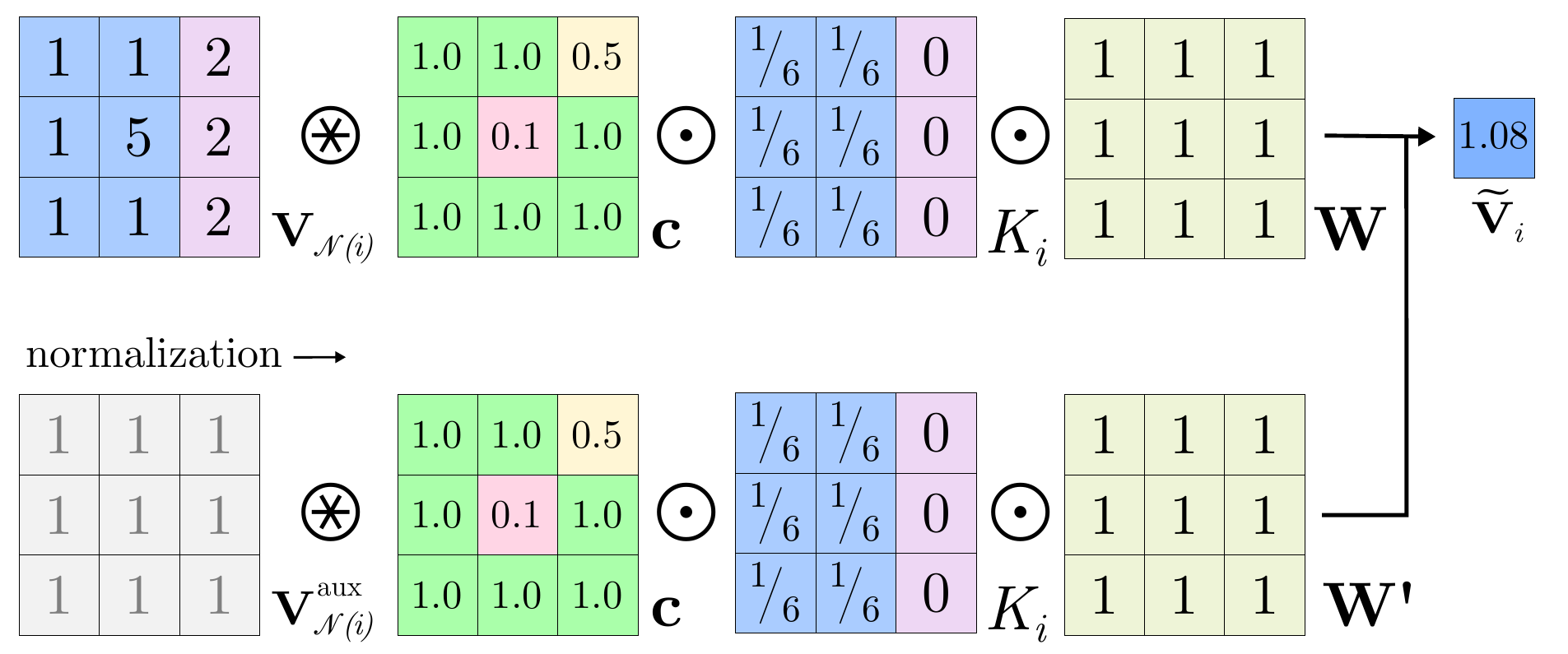}
	\caption{Illustration of our proposed probabilistic PAC approach. The weighting of estimates according to their pixel feature similarity $K$ as well as their confidence $\cc$ allows to remove outliers.}
\label{fig:probabilistic_pac}
\end{figure}
An outlier pixel is surrounded by more reliable estimates, which belong to the same and a different object.
For simplicity, we assume that $\W$ performs an averaging operation.
If the pixel confidence was not taken into account, the outlier value would spread to the surrounding pixels due to its higher magnitude.
In contrast, the proposed PPAC allows to propagate the reliable pixel values from the same object and thus almost completely replaces the outlier in the center.

The normalization as proposed in \cref{sec:advanced_normalization} can be easily extended to PPACs.
To that end, $\tilde{\vv}^{aux}_i$ is obtained as
\begin{equation}
\tilde{\vv}^{aux}_i = \sum_{j \in \mathcal{N}(i)} c_j \cdot K \big(\f_i, \f_j \big) \cdot \W \left[ \p_i - \p_j \right]  \cdot \mathbf{1}.
\label{eq:extended_normalization_pass}
\end{equation}
The normalization step is performed as before and the PPAC outputs are divided per pixel by $\tilde{\vv}^{aux}_i$ before the bias (\Eq~\ref{eq:normalized_pac}) 
\section{Refinement Networks with PPACs}
Deep dense predictors often output predictions at a scale lower than the input resolution to save time and parameters, \eg \cite{Chen:2018:EAS, Yin:2019:HDD}.
The low-resolution estimates are then upscaled by simple methods such as bilinear interpolation.
In \cite{Su:2019:PAC, Wannenwetsch:2019:LTS, Wu:2018:FET2},
image-adaptive convolutions have proven very helpful in this upsampling step.
Following that, we propose a \emph{PPAC refinement network},
which takes image and reliability data into account to upscale and refine network outputs.

A straightforward approach is to upscale results with transposed PPACs.
However, we found that this can lead to difficulties, especially for optical flow.
This is due to the fact that flow networks often assume the input sizes to be divisible by a certain power of 2, \eg $2^6$ \cite{Yin:2019:HDD}.
As this is mostly not the case, \eg for the Sintel benchmark \cite{Butler:2012:NOS},
input images are resized and the output flow is afterwards rescaled with a non-integer factor.
To apply transposed convolutions, which can only upscale by integers,
one has to pad the inputs and crop the output after upscaling.
We observed that this leads to severe artifacts, which clearly reduce the accuracy.

Instead, we propose to first upscale the estimates by the default method of the original network.
A lightweight network with PPACs is then applied at full resolution.
As we only use a small number of PPACs, the computational expense of the approach remains low and the prediction accuracy does not decrease due to padding artifacts or similar.

\begin{figure}[t]
	\centering
	\includegraphics[width=\linewidth]{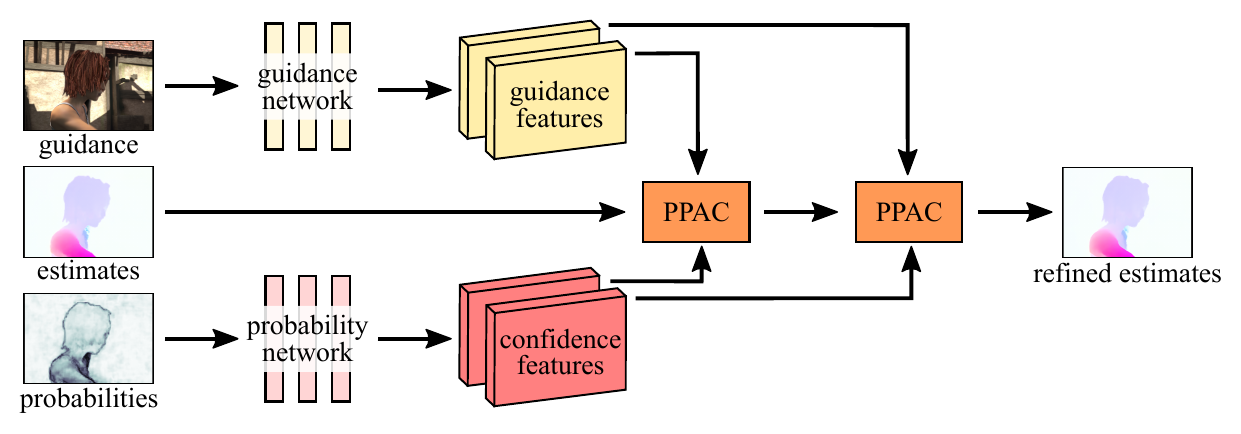}
	\caption{Exemplary architecture of our proposed PPAC refinement networks.}
\label{fig:refinement_network}
\vspace{-0.5em}
\end{figure}

Our proposed refinement networks consist of three branches as illustrated in \cref{fig:refinement_network}.
In addition to the upscaled estimates, the network takes corresponding probability data,
\eg a full marginal posterior per pixel or other probability measures, and the high-resolution images as inputs.
The first subnetwork transforms the probability data into scalar confidence values for the individual PPACs.
The second branch processes the guidance images to generate meaningful pixel features.
Both intermediate outputs as well as the underlying network predictions are then fed into the PPACs of the combination branch to create a refined estimate. 
\section{Implementation}
\subsection{Network architecture}
To illustrate their capabilities, we experiment with PPAC refinement networks for the tasks of optical flow estimation and semantic segmentation.
All networks are fully convolutional and include two consecutive PPACs.
We use log values as inputs to the probability branch and add a sigmoid function at its end for normalization.
Please refer to the supplement for a detailed description of the networks' setup.

Here, we only highlight architectural choices that differ significantly from the ones in \cite{Su:2019:PAC}.
First, we found no benefit from increasing the number of channels in the combination branch.
Moreover, using group convolutions with the number of groups being equal to the number of inputs did not decrease the performance, but significantly lowers the number of parameters.
We even go further and share the convolution weights across all channels, which proved especially beneficial for semantic segmentation, possibly due to the large reduction in parameters.
Following the upscaling setup in \cite{Su:2019:PAC}, we also experimented with standard convolution layers to pre- or post-process the data itself.
However, we found no benefit from such convolutions and thus stick with a combination branch that only includes PPACs.

\subsection{Additional baselines}
\label{sec:additional_baselines}
To assess the benefits of our PPAC refinement networks, we introduce two baseline networks.
The \emph{simple refinement network} takes estimates, log-probabilities, and input images and processes them jointly with standard convolutions.
Here, we set the channel depth such that the number of parameters used for the PPAC network is approximately equal to the simple baseline.
Additionally, we use a PAC baseline that replaces all PPACs with its non-probabilistic PAC counterpart.
For this network, the probability branch is removed and the probabilities are instead concatenated with the guidance images and fed to the guidance subnetwork.
We again ensure that the number of parameters remains comparable for PAC and PPAC refinement networks.

\subsection{Training procedure}
For fair comparison, we only train the refinement networks and do not backpropagate into the original networks themselves.
However, our proposed probabilistic refinement is also easily applicable in fully end-to-end training.
For networks with PACs or PPACs, we apply our advanced normalization from \cref{sec:advanced_normalization}.
Here, we found it important to initialize weights $\W$ and $\W'$ with the same values.
We thus initialize both with positive, random numbers and ensure that $\W'$ remains positive by learning in log-space.

As we aim to compare several different approaches, carrying out ablations on the official test datasets is not feasible.
Thus, we split the data with available ground truth randomly into custom training, validation, and test sets.
Learning rates for the optimization with Adam \cite{Kingma:2015:AAM} are determined for all networks individually on the validation set.
Please see \cref{sec:exps} as well as the supplement for further training specifics.
Our PyTorch code is publicly available.\footnote{\url{https://github.com/visinf/ppac_refinement}} 
\begin{table}[tb]
\centering
  \centering
  \small
  \caption{Average end-point error \emph{(AEE)} and 3-pixel outlier rate \emph{(out)} on our Sintel and KITTI test splits for different normalizations of a PAC refinement network.}
  \label{tab:results_normalization}
  \begin{tabularx}{\linewidth}{@{}XS[table-format=1.3]S[table-format=1.3]S[table-format=1.3]S[table-format=1.2,table-space-text-post={$\,\%$}]@{}}
  \toprule
  & \multicolumn{2}{c}{Sintel (AAE)} &  \multicolumn{2}{@{}c@{}}{KITTI} \\
   \cmidrule(lr){2-3} \cmidrule(l{0.6em}){4-5}
  & {clean} & {final} & {AEE} & {out} \\
  \midrule
  HD3 \cite{Yin:2019:HDD}						& 1.672 					& 1.357 					& 1.990 				& 6.14$\,\%$	\\
  PAC w/o normal.									& 1.665					& 1.352 					& 1.924 				& 6.34$\,\%$ 	\\
  PAC w/ kernel normal.							& 1.622					& 1.323 					& 1.921 				& 6.15$\,\%$ 	\\
  PAC w/ adv.~normal.~\emph{(ours)}	& \bfseries 1.594	& \bfseries 1.302	& \bfseries 1.868 & \bfseries 5.81$\,\%$ \\
  \bottomrule
  \end{tabularx}
  \vspace{-0.5em}
\end{table}

\section{Experiments}
\label{sec:exps}
\subsection{Optical flow}
\label{sec:exp_optical_flow}
We first apply our probabilistic refinement networks to the task of optical flow estimation.
As underlying network, we use the state-of-the-art HD3 method \cite{Yin:2019:HDD}, which yields competitive results on the major benchmarks.
HD3 predicts flow in a residual fashion and estimates a discrete probability distribution at each scale.
In \cite{Yin:2019:HDD}, the full (discrete) matching distribution of the flow is composed, which is time- and memory-consuming.
We instead upsample all probability maps via bilinear interpolation and provide the network with the probability value of the respective flow residual at all five scales.
Since the residuals are subpixel-accurate and mostly fall outside of the discrete grid, we use a nearest neighbor interpolation of the probabilities.

We evaluate the proposed refinement networks on two widely used optical flow benchmarks:
Sintel \cite{Butler:2012:NOS} and KITTI \cite{Geiger:2012:AWR2, Menze:2015:OSF}.
The Sintel data is split into 862 images for training, 80 for validation, and 99 for test.
Moreover, we merge the KITTI 2012 and 2015 images to obtain 319 training, 31 validation, and 44 test images.
Using the procedure described in \cite{Yin:2019:HDD}, we fine-tune two individual HD3 models on our own Sintel and KITTI training splits.
We initialize the networks from the author-provided checkpoint pre-trained on FlyingChairs \cite{Dosovitskiy:2015:FNL} as well as FlyingThings3D \cite{Ilg:2017:FNE} and determine the best models on our validation images.
Following \cite{Yin:2019:HDD}, only images from Sintel final are used for fine-tuning of the Sintel model.

All refinement networks are trained for 500 epochs with the average end-point error (AEE) as loss function and a batch size of 8.
The base learning rates are cut by a factor of two every $100$ epochs.
As augmentations, we only apply random cropping to sizes $(384, 768)$ and $(320, 896)$ for Sintel and KITTI data, respectively.
On Sintel, individual networks are trained on the final and clean subsets.

\myparagraph{Comparison of normalization approaches.}
We first demonstrate the benefit of our proposed advanced normalization scheme.
Therefore, we train PAC refinement networks without normalization, similar to \cite{Su:2019:PAC}, and with kernel normalization, \cf \cref{sec:advanced_normalization}.
\cref{tab:results_normalization} shows results evaluated on our test sets of Sintel and KITTI.
Note that the results on Sintel clean tend to be worse than on final as this data has not been seen during HD3 fine-tuning.
We observe that a PAC network with kernel normalization is able to improve the accuracy over the HD3 baseline as well as the PAC approach without normalization.
However, the same network leveraging our proposed advanced normalization shows clearly the best accuracy on all test sets.
We attribute this to the fact that kernel normalization does not sufficiently compensate different neighborhood conditions.

\myparagraph{Evaluation of PPACs.}
We now evaluate refinement networks including PPACs in comparison to the simple and PAC baselines as introduced in \cref{sec:additional_baselines}.
The results of flow refinement on our test splits are summarized in \cref{tab:results_flow}.
\begin{table}[tb]
\centering
  \centering
  \small
  \caption{Average-end-point error \emph{(AEE)} and 3-pixel outlier rate \emph{(out)} on our Sintel and KITTI test splits for different refinements.}
  \label{tab:results_flow}
  \begin{tabularx}{\linewidth}{@{}XS[table-format=1.3]S[table-format=1.3]c<{\vphantom{a}}S[table-format=1.3]S[table-format=1.2,table-space-text-post={$\,\%$}]@{}}
  \toprule
  & \multicolumn{2}{@{}c@{}}{Sintel (AAE)} && \multicolumn{2}{@{}c@{}}{KITTI}  \\
   \cmidrule(lr){2-3} \cmidrule(l{0.6em}){5-6}
  & {clean} & {final} && {AEE} & {out} \\
  \midrule
  HD3 \cite{Yin:2019:HDD}	& 1.672 					& 1.357 					&& 1.990 					& 6.14$\,\%$	\\
  Simple refinement					& 1.638					& 1.334 					&& 1.872 					& 5.92$\,\%$ \\
  PAC network	\cite{Su:2019:PAC}	& 1.594					& 1.302					&& 1.868 					& 5.81$\,\%$	\\
  PPAC network \emph{(ours)}			& \bfseries 1.562	& \bfseries 1.283	&& \bfseries 1.848	& \bfseries 5.50$\,\%$	 \\
  Oracle network					& \textit{1.430}		&  \textit{1.149}		&& \textit{1.500}		& \textit{4.48}$\,\%$	\\
  \midrule
 SL \cite{Wannenwetsch:2019:LTS}	& 1.634	& 1.340 	&& 1.953	& 6.41$\,\%$	\\
 FBS \cite{Barron:2016:FBS} 				& 1.643	& 1.354	&& {--}		& {--} \\
  \bottomrule
  \end{tabularx}
  \vspace{-0.5em}
\end{table}
The simple refinement approach based only on standard convolutions (with the same inputs) improves the flow predictions only slightly on all datasets.
Applying guidance data, including the estimated probabilities, explicitly in a PAC refinement network already leads to a clear improvement.
However, our proposed PPAC refinement network outperforms the PAC approach by a large margin, improving the AEE by $6.6\%$, $5.5\%$ and $7.1\%$ on Sintel clean, Sintel final and KITTI, respectively.
We also observe that the improvement of using content-adaptive, probabilistic convolutions over a simple setup with standard convolutions is more significant on Sintel than on KITTI.
We attribute this to the fact that guidance data has shown to be most effective in boundary regions \cite{Wannenwetsch:2019:LTS}, which play a lesser role in the KITTI dataset as the ground truth is sparse.

\begin{figure}[t]
  \begin{subfigure}[b]{0.49\linewidth}
  \centering
  \includegraphics[width=\linewidth]{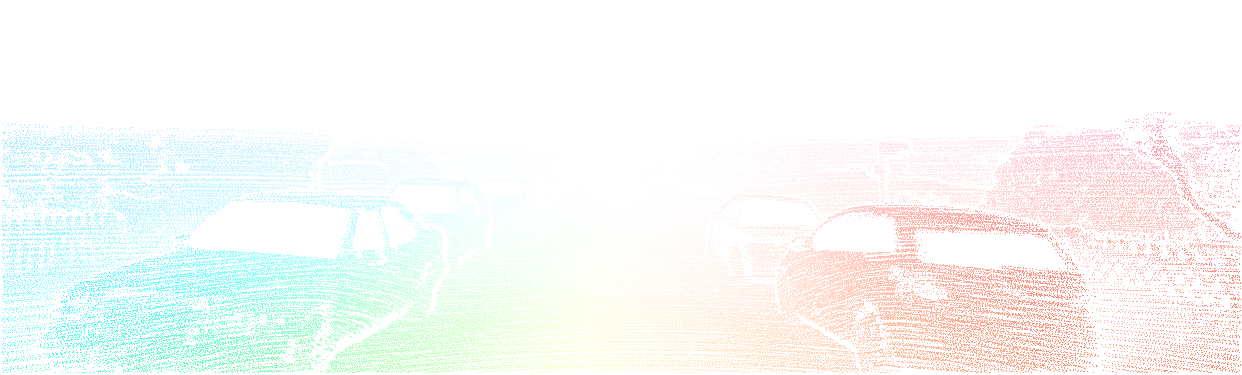}
  \end{subfigure}
  \hfill
  \begin{subfigure}[b]{0.49\linewidth}
    \centering
    \includegraphics[width=\linewidth]{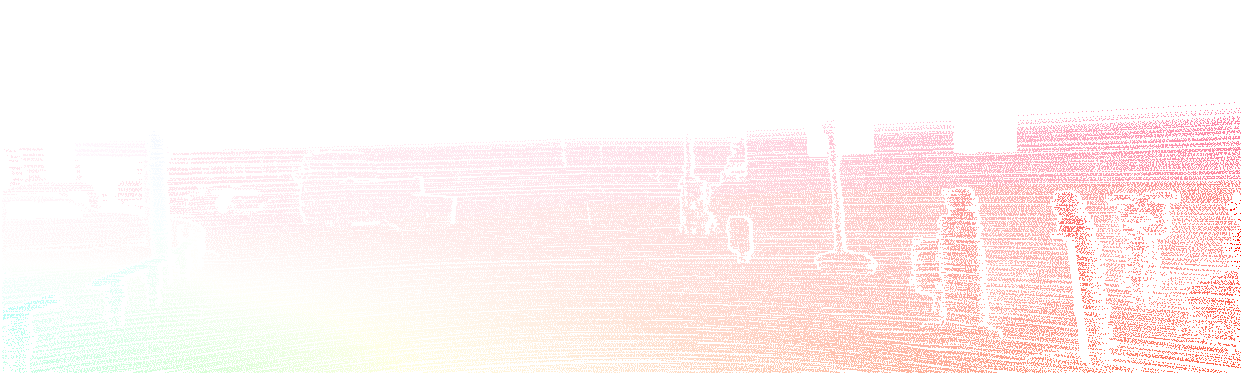}
  \end{subfigure} \\
  \begin{subfigure}[b]{0.49\linewidth}
    \centering
    \includegraphics[width=\linewidth]{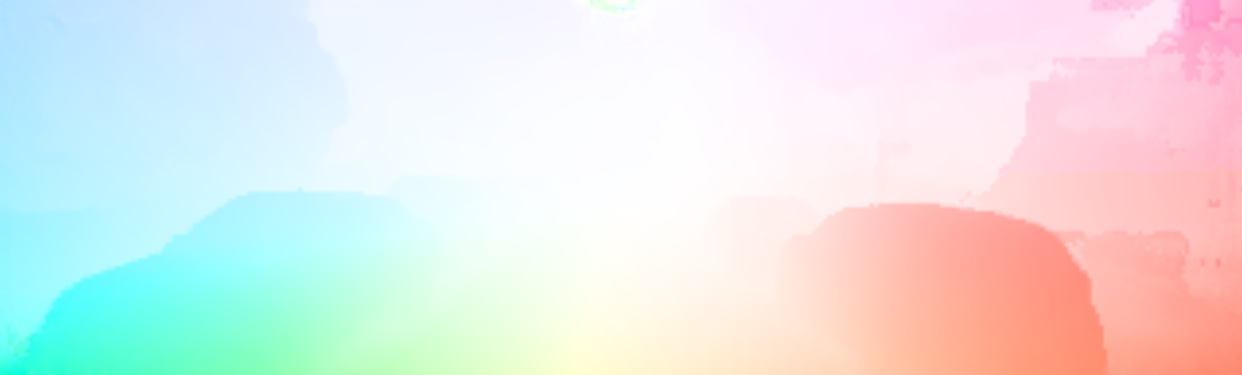}
  \end{subfigure}
  \hfill
  \begin{subfigure}[b]{0.49\linewidth}
    \centering
    \includegraphics[width=\linewidth]{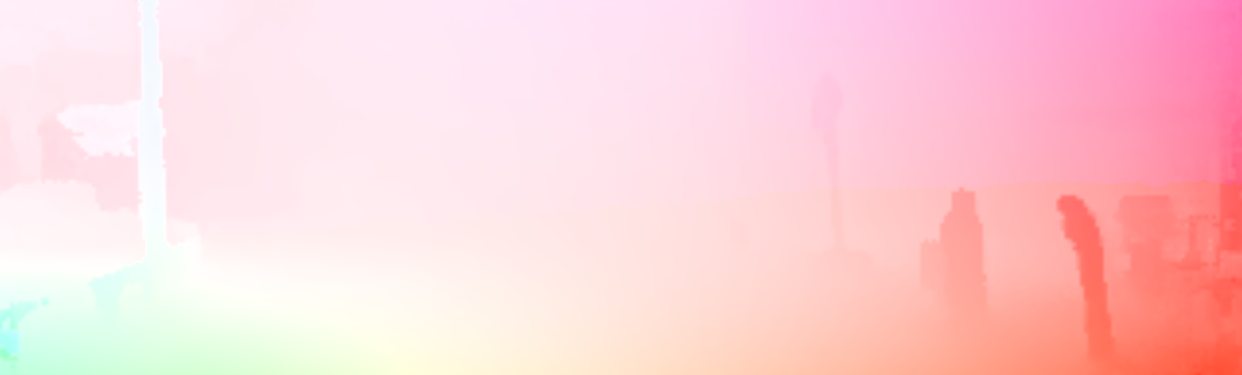}
	\end{subfigure} \\
  \begin{subfigure}[b]{0.49\linewidth}
    \centering
    \includegraphics[width=\linewidth]{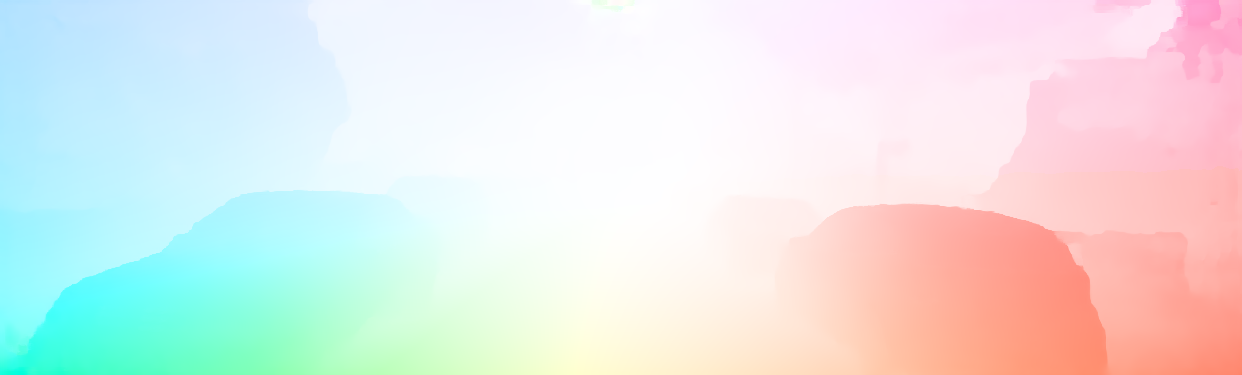}
  \end{subfigure}
  \hfill
  \begin{subfigure}[b]{0.49\linewidth}
    \centering
    \includegraphics[width=\linewidth]{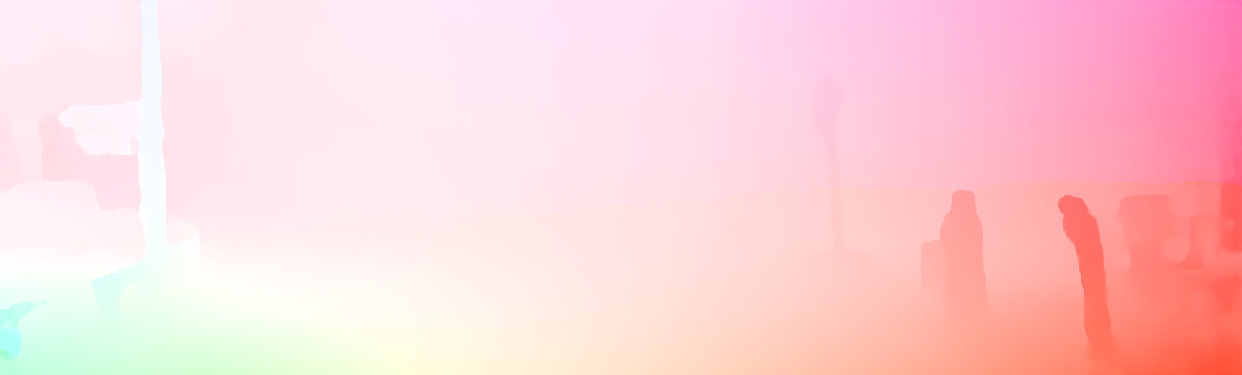}
	\end{subfigure}
  \caption{Examples of ground truth \emph{(top)}, HD3 optical flow \emph{(middle)}, and PPAC-refined optical flow \emph{(bottom)} on KITTI. 
  \emph{Best viewed on screen.}}
  \label{fig:examples_kitti}
  \vspace{-0.5em}
\end{figure}

Even more striking than these significant accuracy gains, are the improvements in visual quality.
\cref{fig:figure1,fig:examples_kitti} show example flow fields on Sintel final and KITTI.
PPACs clearly improve the underlying HD3 estimates and lead to substantial improvements especially near motion boundaries.
Additionally, our proposed approach is able to correctly propagate flow estimates into outlier regions.

To further understand the results of PPAC refinement, we evaluate an additional \emph{oracle network}, which takes oracle confidences as input to the probability branch, which we take to be the inverse of the AEE.
As this correctly assesses the reliability of each pixel, this networks provides an upper bound on the possible accuracy improvement from probabilistic refinement.
Comparing the results in \cref{tab:results_flow}, we observe that an even more significant improvement would be possible if more precise probability estimates were available.
This observation holds especially for the evaluation on KITTI, where a rather small amount of ground truth is available during fine-tuning.
This suggests that future work on improved probability estimates in deep network has the potential of improving the accuracy in difficult areas further.

We finally compare our proposed PPAC refinement to other approaches from the literature.
As PACs have shown to be the state-of-the-art method for joint upsampling,
we restrict further comparison to the Semantic Lattice (SL) \cite{Wannenwetsch:2019:LTS}, which appeared concurrently to \cite{Su:2019:PAC}, and the Fast Bilateral Solver (FBS) \cite{Barron:2016:FBS} as a representative method that explicitly considers confidences for post-processing.
The SL is trained as described in \cite{Wannenwetsch:2019:LTS}.
For the FBS, we use the probabilities of the last output layer of HD3 \cite{Yin:2019:HDD}.
All FBS weight parameters and the trade-off parameter $\lambda$ are determined via grid search on the validation set.
Note that we were not able to find stable parameters for FBS on KITTI.
\cref{tab:results_flow} shows that SL and FBS are both able to improve the HD3 baseline accuracy.
However, PPAC refinement outperforms both previous methods by a large margin.

\myparagraph{Evaluation on benchmarks.}
We finish our flow experiments by evaluating PPAC-HD3, \ie HD3 optical flow \cite{Yin:2019:HDD} with PPAC refinement, on the official benchmarks.
For Sintel, we initialize HD3 with the same checkpoint as used in \cite{Yin:2019:HDD}.
On KITTI, we use the fine-tuned checkpoint with context module as provided online.
To train our PPAC network, we leverage the entire training data provided for the Sintel and KITTI benchmarks, respectively.
Again, we train separate networks for Sintel clean and final, as well as a joint refinement network for both KITTI benchmarks.
In comparison to our previous experiment, we adjust the number of epochs in the learning scheme such that the total number of iterations remains approximately the same despite the larger number of images.
Note that we do not use other augmentations than random cropping, since we found our refinement network to be robust \wrt overfitting.
We attribute this to the lightweight structure of our PPAC network, which only adds approximately $12$k parameters.

Results on Sintel test are summarized in \cref{tab:results_flow_sintel_test}.
\begin{table}[tb]
\centering
  \centering
  \small
  \caption{Average-end-point error \emph{(AEE)} and AEE of regions $\leq$10 pixels from occlusion boundaries \emph{(d0-10)} evaluated on Sintel test.}
  \label{tab:results_flow_sintel_test}
  \begin{tabularx}{\linewidth}{@{}XS[table-format=1.3]S[table-format=1.3]c<{\vphantom{a}}S[table-format=1.3]S[table-format=1.3]@{}}
  \toprule
  & \multicolumn{2}{c}{clean} && \multicolumn{2}{c@{}}{final}  \\
   \cmidrule(lr){2-3} \cmidrule(l){5-6}
  & {AEE} & {d0-10} && {AEE} & {d0-10} \\
  \midrule
  HD3 \cite{Yin:2019:HDD}	& 4.788 					& 	3.225					&& 4.666					& 3.786 \\
  PPAC-HD3 \emph{(ours)}	& \bfseries 4.589	&  \bfseries	2.788 && \bfseries 4.599	& \bfseries 3.521 \\
  \bottomrule
  \end{tabularx}
\end{table}
PPAC-HD3 clearly improves the accuracy of the underlying HD3 method on clean and final splits by $\sim$4.2$\%$ and $\sim$1.4$\%$, respectively.
The larger improvement on clean might be partly due to the fact that no data from this pass was used during fine-tuning of the HD3 baseline.
Moreover, we observe a very significant improvement of $7.0\%$ on the final pass and of $\sim$13.6$\%$ on Sintel clean when considering the AEE of regions closer than 10 pixels to motion boundaries \emph{(d0-10)}.
Overall, PPAC-HD3 ranks \nth{4} on Sintel final among published two-frame methods and is thus highly competitive.

In \cref{tab:results_flow_kitti_test}, we show results on the official test sets of KITTI 2012 and 2015.
\begin{table}[tb]
\centering
  \centering
  \small
  \caption{3-pixel outlier rate of non-occluded/all pixels \emph{(Out-Noc/all)} and runtimes evaluated on KITTI 2012 and 2015 test.}
  \label{tab:results_flow_kitti_test}
  \begin{tabularx}{\linewidth}{@{}XS[table-format=1.2,table-space-text-post={$\,\%$}]S[table-format=1.2,table-space-text-post={$\,\%$}]S[table-format=1.2,table-space-text-post={$\,\%$}]S[table-format=1.2,table-space-text-post=\,s]@{}}
  \toprule
  & \multicolumn{2}{c}{KITTI 2012} & \multicolumn{2}{c@{}}{KITTI 2015}  \\
   \cmidrule(l{0.6em}r{0.6em}){2-3} \cmidrule(l){4-5}
  & {Out-Noc} & {Out-all} & {Out-all} & {Runtime} \\
  \midrule
  HD3 \cite{Yin:2019:HDD}	& 2.26$\,\%$					& 	5.41$\,\%$					& 6.55$\,\%$					& \bfseries 0.11\,s \\
  PPAC-HD3 \emph{(ours)}	& \bfseries 2.01$\,\%$	& \bfseries	5.09$\,\%$	& \bfseries 6.06$\,\%$		& 0.19\,s \\
  \bottomrule
  \end{tabularx}
  \vspace{-0.5em}
\end{table}
PPAC-HD3 outperforms its underlying method by a large margin, leading to a substantial relative improvement of $\sim$11.1$\%$ for the outlier rate of non-occluded pixels on KITTI 2012.
On KITTI 2015, PPAC refinement improves the outlier rate of all pixels from $6.55\%$  to $6.06\%$ (relative improvement of $\sim$7.5$\%$).
Our proposed method clearly ranks \nth{1} among published optical flow approaches on both datasets.
PPAC-HD3 is even able to outperform several strong scene flow methods, which leverage additional stereo data as input.
\cref{tab:results_flow_kitti_test} also shows computational times measured on a single GTX 1080 Ti GPU.
Adding PPACs has a computational overhead of $\sim 1.7\times$, which seems justified by the strongly improved results.
 
\subsection{Semantic segmentation}
In a second set of experiments, we apply our probabilistic refinement networks to the task of semantic segmentation.
We choose DeepLabv3+ \cite{Chen:2018:EAS} as a baseline using the XCeption65 variant of the model.
Before feeding the logits of the segmentation network into the probability branch of our PPAC network, we normalize them with a log-softmax operation.
The logits are equally used as input to the combination branch.
Here, we left the values unnormalized as we found a normalization to lead to inferior results.

We evaluate our probabilistic refinement network on Pascal VOC 2012 \cite{Everingham:2015:PVO} and use the training, validation, and test split of \cite{Wannenwetsch:2019:LTS}.
When training PAC and PPAC networks, we found it important to use different learning rates for the preprocessing branches as well as the weights in the combination branch.
To determine appropriate values, we first fixed the PPAC or PAC weights to a Gaussian kernel with standard deviation $\sigma=1$ and searched for the optimal learning rate of the guidance and probability subnetworks on the validation set.
In a second step, all weights are randomly initialized and a second learning rate is determined for the convolution weights in the combination branch.
Furthermore, we observed improved accuracy when initializing the bias term of PACs and PPACs to zero.
All networks are trained for $500$ epochs with constant learning rate, a batchsize of $16$, and random image crops with size $200 \times 272$.
We use the cross entropy as loss function and evaluate the mean intersection over union for validation.
Please see the supplement for details, \eg the network architectures.

\cref{tab:results_segmentation} summarizes the results on our test split of Pascal VOC 2012.
\begin{table}[tb]
  \centering
  \small
  \caption{Mean intersection over union \emph{(mIoU)} evaluated on our Pascal VOC 2012 test split. $^{\dag}$Results taken from \cite{Wannenwetsch:2019:LTS}.}
  \label{tab:results_segmentation}
  \begin{tabularx}{\linewidth}{@{}XS[table-format=2.2,table-space-text-post={$\,\%$}]@{}}
  \toprule
    & {mIoU} \\
  \midrule
  DeepLabv3+ \cite{Chen:2018:EAS} & 82.20$\,\%$ 	\\
  PAC refinement \cite{Su:2019:PAC}		& 82.39$\,\%$ \\
  PPAC refinement	\emph{(ours)} & \bfseries 82.62$\,\%$ \\
  \midrule
  SL \cite{Wannenwetsch:2019:LTS}$^{\dag}$	& 82.25$\,\%$ \\
   FBS \cite{Barron:2016:FBS}				& 82.28$\,\%$	\\
  \bottomrule
 \end{tabularx}
 \vspace{-0.5em}
 \end{table}
In this setting, our PPAC network requires on average 0.055s per image on a single GTX 1080 Ti GPU and is able to improve the segmentation accuracy
even though DeepLabv3+ takes already special care of the decoder.
In contrast, PAC refinement shows a considerably smaller benefit.
We were not able to find a simple configuration that improved the original results.
As for optical flow, we also compare to SL \cite{Wannenwetsch:2019:LTS} and FBS \cite{Barron:2016:FBS}, and use the probability of the most likely class as confidence input.
Both previous methods are clearly outperformed by our proposed PPACs.

\begin{figure}[t]
\centering
  \begin{subfigure}[b]{0.49\linewidth}
  \centering
  \includegraphics[width=\linewidth]{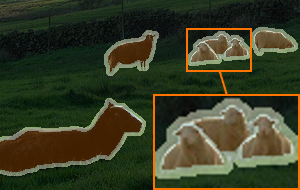}
  \end{subfigure}
  \hfill
  \begin{subfigure}[b]{0.49\linewidth}
    \centering
    \includegraphics[width=\linewidth]{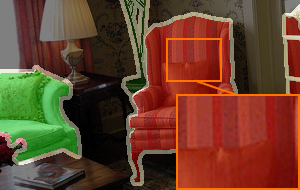}
  \end{subfigure} \\
  \vspace{0.1em}
  \begin{subfigure}[b]{0.49\linewidth}
    \centering
    \includegraphics[width=\linewidth]{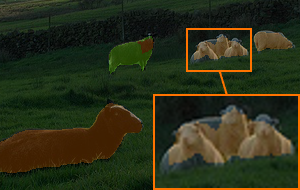}
  \end{subfigure}
  \hfill
  \begin{subfigure}[b]{0.49\linewidth}
    \centering
    \includegraphics[width=\linewidth]{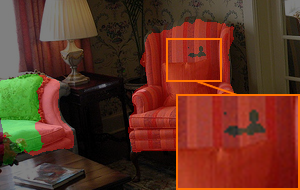}
	\end{subfigure} \\
	\vspace{0.1em}
  \begin{subfigure}[b]{0.49\linewidth}
    \centering
    \includegraphics[width=\linewidth]{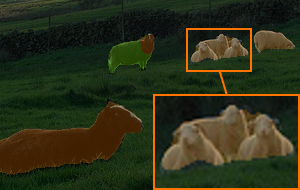}
  \end{subfigure}
  \hfill
  \begin{subfigure}[b]{0.49\linewidth}
    \centering
    \includegraphics[width=\linewidth]{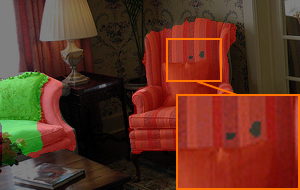}
	\end{subfigure}
  \caption{Cropped examples of ground truth \emph{(top)}, DeepLabv3+ \emph{(middle)}, and PPAC refined segmentation maps \emph{(bottom)} on Pascal VOC 2012. \emph{Best viewed on screen.}}
  \label{fig:examples_pascal}
   \vspace{-0.5em}
\end{figure}

\cref{fig:examples_pascal} shows examples of segmentation maps from Pascal VOC 2012.
PPAC refinement leads to a better alignment with the underlying objects especially at object intersections or for smaller objects.
Moreover, PPACs are able to successfully close smaller holes in the segmentation maps. 
\section{Conclusion}
We introduced probabilistic pixel-adaptive convolutions (PPACs), which allow for filtering operations that respect guidance data as well as per-pixel confidences.
Building on the work of \cite{Su:2019:PAC}, we first proposed an advanced normalization scheme, which we show to clearly improve the results in practice.
Subsequently, we extend PACs to include confidence information during the filtering step to especially improve regions of low reliability.
We proposed to use PPACs for the refinement of dense prediction networks and demonstrated their benefits for optical flow estimation and semantic segmentation.
Here, PPAC refinement resulted in significant accuracy gains; our PPAC-HD3 clearly leads both KITTI benchmarks for optical flow.
Moreover, refined estimates show fewer boundary artifacts and are smoother overall while correctly respecting object boundaries. 
{\small
\bibliographystyle{ieee_fullname}
\bibliography{short,lit,external,papers}
}
\flushcolsend

\title{Probabilistic Pixel-Adaptive Refinement Networks \\ {\large -- Supplemental Material --}}
\author{Anne S. Wannenwetsch$^{1,2*}$ \qquad Stefan Roth$^2$ \\
$^1$Amazon, Germany \qquad $^2$ TU Darmstadt, Germany}

\maketitle

In this supplemental material, we give further implementation details of the different types of refinement networks and provide results for a more comprehensive comparison on optical flow benchmarks.
Moreover, we present an analysis considering the PPAC improvements on unreliable pixels as well as additional visualizations of PPAC-refined optical flow fields and segmentation maps.

\appendix

\section{Additional Implementation Details}
\subsection{Learning procedure}
To train our networks, we use the Adam optimizer \cite{Kingma:2015:AAM} with default parameters $\beta_1=0.9$, $\beta_2=0.999$ and without weight decay.
PPAC refinement networks are trained with a learning rate of \num{1e-3} for networks on Sintel and a learning rate of \num{5e-3} on KITTI.
For semantic segmentation on Pascal VOC 2012, we use a learning rate of \num{1e-4} for guidance and probability branches and \num{1e-5} for the remaining PPAC parameters.
The image inputs to all networks are normalized while estimates and log-probabilities remain unchanged.
For faster training of all refinement networks, we save the outputs of the underlying backbone networks (\ie HD3 or DeepLabv3+), and only propagate through the refinement step.

\subsection{Network architectures}
\cref{tab:arch_ppac,tab:arch_pac,tab:arch_simple} show the network structures used for PPAC, PAC, and our baseline simple refinement network, respectively.
\begin{table}[tb]
  \centering
  \small
  \caption{Network structure of our PPAC networks for optical flow/semantic segmentation with $\sim$12.3k/14.3k parameters.}
  \label{tab:arch_ppac}
  \begin{tabularx}{\linewidth}{@{}Xr@{\,}c@{\,}lccc@{}}
  \toprule
  & \multicolumn{3}{c}{layer} & kernel & non- & shared \\
  & \multicolumn{3}{c}{type} & size & linearity & weights \\
  \midrule
  guidance	& C: $3$&$\rightarrow$&$15$	& $5\times5$ & ReLU 		& \xmark	\\
  branch		& C: $15$&$\rightarrow$&$15$	& $5\times5$ & ReLU 		& \xmark	\\
  					& C: $15$&$\rightarrow$&$10$	& $5\times5$ & \xmark	& \xmark	\\
  \midrule
  probability	& C: $5/21$&$\rightarrow$&$5$& $5\times5$ & ReLU 	& \xmark	\\
  branch		& C: $5$&$\rightarrow$&$5$		& $5\times5$ & ReLU 		& \xmark	\\
  					& C: $5$&$\rightarrow$&$2$		& $5\times5$ & Sigmoid	& \xmark	\\
  \midrule
  combination	& PP: $2/21$&$\rightarrow$&$2/21$	& $7\times7$ & \xmark		& \cmark	\\
  branch			& PP: $2/21$&$\rightarrow$&$2/21$	& $7\times7$ & \xmark		& \cmark	\\
  \bottomrule
  \end{tabularx}
\end{table}
\begin{table}[tb]
  \centering
  \small
  \caption{Network structure of our PAC baseline networks for optical flow/semantic segmentation with $\sim$12.6k/15.5k parameters.}
  \label{tab:arch_pac}
  \begin{tabularx}{\linewidth}{@{}Xr@{\,}c@{\,}lccc@{}}
  \toprule
  & \multicolumn{3}{c}{layer} & kernel & non- & shared \\
  & \multicolumn{3}{c}{type} & size & linearity & weights \\
  \midrule
  guidance	& C: $8/24$&$\rightarrow$&$15/13$	& $5\times5$ & ReLU 		& \xmark	\\
  branch		& C: $15/13$&$\rightarrow$&$15/13$	& $5\times5$ & ReLU 		& \xmark	\\
  					& C: $15/13$&$\rightarrow$&$10$		& $5\times5$ & \xmark	& \xmark	\\
  \midrule
  combination	& P: $2/21$&$\rightarrow$&$2/21$	& $7\times7$ & \xmark		& \cmark	\\
  branch			& P: $2/21$&$\rightarrow$&$2/21$	& $7\times7$ & \xmark		& \cmark	\\
  \bottomrule
  \end{tabularx}
\end{table}
 \begin{table}[tb]
  \centering
  \small
  \caption{Network structure of our simple baseline network for optical flow with a total of $\sim$12.4k parameters.}
  \label{tab:arch_simple}
  \begin{tabularx}{\linewidth}{@{}Xr@{\,}c@{\,}lccc@{}}
  \toprule
  & \multicolumn{3}{c}{layer} & kernel & non- & shared \\
  & \multicolumn{3}{c}{type} & size & linearity & weights \\
  \midrule
  simple									& C: $10$&$\rightarrow$&$11$	& $7\times7$ & ReLU		& \xmark	\\
  branch									& C: $11$&$\rightarrow$&$11$	& $7\times7$ & ReLU		& \xmark	\\
  												& C: $11$&$\rightarrow$&$2$	& $7\times7$ & \xmark	& \xmark	\\
  \bottomrule
  \end{tabularx}
\end{table}
Here, `C' represents standard convolution layers, `P' layers with non-probabilistic PACs, and `PP' layers with our PPACs.
The networks for optical flow and semantic segmentation differ mainly by the number of input and output channels ($2$ or $21$, respectively).
For optical flow, the guidance branch uses only the first image as input since the flow fields should be aligned \wrt the objects in this image.
All standard as well as PAC and PPAC-convolutions pad the inputs with zeros to preserve the feature size and use a stride of one.
Moreover, a bias term is learned for all types of convolutions.
The output of guidance and, if applicable, probability branches is split equally by the number of PAC or PPACs such that individual guidance is used for the components of the combination branch.
For the simple setup, we equally experimented with networks with two convolutions and thus more channels but found the given one with three convolutions to perform better. \section{Detailed Comparison on Optical Flow Benchmarks}
For completeness, we give a more detailed comparison on benchmarks for optical flow, including the training results of PPAC-HD3 as well as the results of related work.

\cref{tab:benchmark_sintel} shows results on Sintel clean and final.
\begin{table}[tb]
\centering
  \centering
  \small
  \caption{Average end-point error \emph{(AEE)} of top-ranked two-frame optical flow methods on Sintel train and test. $^{\star}$Re-evaluated for comparability.}
  \label{tab:benchmark_sintel}
  \begin{tabularx}{\linewidth}{@{}XS[table-format=1.2,table-space-text-pre={(},table-space-text-post={$)^\star$}]S[table-format=1.2,table-space-text-pre={(},table-space-text-post={$)^\star$}]S[table-format=1.2]S[table-format=1.2]@{}}
  \toprule
  &  \multicolumn{2}{c}{train} & \multicolumn{2}{c}{test} \\
  \cmidrule(lr){2-3} \cmidrule(l){4-5}
  & {clean} & {final} & {clean} & {final} \\
  \midrule
  VCN \citesupp{Yang:2019:VCN} & (1.66) & (2.24) & \bfseries 2.81 & \bfseries 4.40 \\
  IRR-PWC \cite{Hur:2019:IRR2} & (1.92) & (2.51) & 3.84  & 4.58  \\
  PWC-Net+ \citesupp{Sun:2018:MMS} & (1.71) & (2.34) & 3.45 & 4.60  \\
  PPAC-HD3 \emph{(ours)} & \bfseries (1.54) & \bfseries (1.05) & 4.59 & 4.60 \\
  HD3 \cite{Yin:2019:HDD} & (1.68)$^{\star}$ & (1.15)$^{\star}$ & 4.79 & 4.67 \\
  \bottomrule
  \end{tabularx}
\end{table}
For comparability, we re-evaluated the flow fields of HD3 on the training splits, taking into account the available invalid masks.
Our proposed PPAC-HD3 ranks \nth{4} \wrt to the AEE on Sintel final.

\begin{table}[tb]
\centering
  \centering
  \small
  \caption{Average end-point error \emph{(AEE)} and 3-pixel outlier rate on non-occluded/all pixels \emph{(Out-Noc/all)} of top-ranked optical flow methods on KITTI 2012 train and test. Results in parentheses indicate that data was used in training. $^{\dag}$Methods use left and right stereo images. $^{\star}$Re-evaluated for comparability.}
  \label{tab:benchmark_kitti12}
  \begin{tabularx}{\linewidth}{@{}XS[table-format=1.2,table-space-text-pre={(},table-space-text-post={$)^\star$}]S[table-format=1.1]S[table-format=1.2,table-space-text-post={$\,\%$}]S[table-format=1.2,table-space-text-post={$\,\%$}]@{}}
  \toprule
  &  {train} & \multicolumn{3}{c}{test} \\
  \cmidrule(lr){2-2} \cmidrule(l){3-5}
  & {AEE} & {AEE} & {Out-Noc} & {Out-all} \\
  \midrule
  PPAC-HD3 \emph{(ours)} & \bfseries (0.71) & 1.2 & \bfseries 2.01$\,\%$ & 5.09$\,\%$  \\
  HD3 \cite{Yin:2019:HDD} & (0.81)$^{\star}$ & 1.4 & 2.26$\,\%$ &  5.41$\,\%$ \\
  PRSM$^{\dag}$ \citesupp{Vogel:2015:3SF} & {--} & \bfseries 1.0 & 2.46$\,\%$ & \bfseries 4.23$\,\%$ \\
  LiteFlowNet2 \citesupp{Hui:2019:LOF} & {--} & 1.4 & 2.63$\,\%$ & 6.16$\,\%$ \\
  SPS-StFl$^{\dag}$ \citesupp{Yamaguchi:2014:EJS} & {--} & 1.3 & 2.82$\,\%$ & 5.61$\,\%$ \\
  \bottomrule
  \end{tabularx}
\end{table}

\begin{table}[tb]
\centering
  \centering
  \small
  \caption{Average end-point error \emph{(AEE)}, 3-pixel outlier rate on all pixels \emph{(Out-all)}, and runtimes \emph{(time)} of top-ranked methods on KITTI 2015 train and test. Results in parentheses indicate that data was used in training. $^{\dag}$Methods use left and right stereo images. $^{\star}$Re-evaluated for comparability.}
  \label{tab:benchmark_kitti15}
  \begin{tabularx}{\linewidth}{@{}XS[table-format=1.2,table-space-text-pre={(},table-space-text-post={$)^\star$}]@{\hspace{0.5em}}S[table-format=1.2,table-space-text-pre={(},table-align-text-post=true,table-space-text-post={$\,\%)^\star$}]@{\hspace{1em}}S[table-format=1.2,table-space-text-post={$\,\%$}]@{\hspace{0.5em}}S[table-format=3.2,table-align-text-post=true,table-space-text-post={\,s$^\star$}]@{}}
  \toprule
  &  \multicolumn{2}{c}{train} & \multicolumn{2}{c}{test} \\
  \cmidrule(lr{1em}){2-3} \cmidrule{4-5}
  &  {AEE} & {Out-all} & {Out-all} & {time} \\
  \midrule
 UberATG-DRISF$^{\dag}$ \citesupp{Ma:2019:DRI} & {--} & {--} & \bfseries 4.73$\,\%$ & 0.75\,s \\
 PPAC-HD3 \emph{(ours)} & (1.20) & \bfseries (3.56$\,\%$) & 6.06$\,\%$ & 0.19\,s \\
 ISF$^{\dag}$ \citesupp{Behl:2017:BBS} & {--} & {--} & 6.22$\,\%$ & 600\,s \\
 VCN \citesupp{Yang:2019:VCN} & \bfseries (1.16) & (4.1$\,\%$) & 6.30$\,\%$ & 0.18\,s \\
  HD3 \cite{Yin:2019:HDD} & (1.40)$^{\star}$ & (4.39$\,\%$)$^{\star}$ & 6.55$\,\%$ & \bfseries 0.11\,s$^{\star}$ \\
  \bottomrule
  \end{tabularx}
\end{table}
\cref{tab:benchmark_kitti12,tab:benchmark_kitti15} summarize results for the best-ranked published methods on KITTI 2012 and 2015.
For completeness, we also include scene flow methods. 
Note, however, that such approaches are not fully comparable as they leverage additional stereo images to compute flow.
On both datasets, PPAC-HD3 ranks \nth{1} among optical flow methods and \nth{2} over all published approaches on KITTI 2015.
As we used the publically available checkpoint for HD3, which differs slightly from the one used in \cite{Yin:2019:HDD}, we report re-evaluated results on the training sets.
Moreover, we provide HD3 runtimes evaluated on the same GTX 1080 Ti GPU as PPAC-HD3 for fair comparison.

 \section{Improvement of Unreliable Pixels}
In the main paper, we argue that probabilistic pixel-adaptive refinement allows to propagate correct estimates into unreliable regions.
Here, we examine the influence of PPACs on unreliable pixels in more detail.
We evaluate the refinement of optical flow by computing the AEE on the $10\%$ most unreliable pixels of each flow field and comparing it to the AEE of the remaining pixels.
To assess the reliability of a pixel estimate, we upsample the probabilities of the last output scale and use nearest neighbor interpolation if the estimated residuals fall outside the probability grid.
We found these reliabilities to correlate better with the optical flow errors than the ones obtained from the composed full matching probability distribution proposed in \cite{Yin:2019:HDD}.
Moreover, we use the same PPAC refinement networks as trained for the experiments in \cref{tab:results_flow}.

\cref{tab:improvement_reliability} shows the relative improvement on unreliable and remaining pixels evaluated on our test splits of Sintel and KITTI.
\begin{table}[tb]
\centering
  \centering
  \small
  \caption{Relative improvement of average end-point error \emph{(AEE)}, evaluated on the $10\%$ most unreliable and the remaining pixels of our Sintel and KITTI test splits.}
  \label{tab:improvement_reliability}
  \begin{tabularx}{\linewidth}{@{}XS[table-format=1.2,table-space-text-post={$\,\%$}]S[table-format=1.2,table-space-text-post={$\,\%$}]S[table-format=1.2,table-space-text-post={$\,\%$}]@{}}
  \toprule
  & \multicolumn{2}{c}{Sintel} &  \multicolumn{1}{c@{}}{KITTI} \\
   \cmidrule(lr){2-3} 
  & {clean} & {final} & \\
  \midrule
  Most unreliable pixels	& 9.86\,\%	& 8.93\,\%	& 4.28\,\%	\\
  Remaining pixels 		& 4.11\,\%	& 3.01\,\%	& 9.42\,\%	\\
  \bottomrule
  \end{tabularx}
\end{table}
On Sintel clean and final, we clearly observe a more significant improvement on the unreliable pixels, justifying the conclusion that PPACs allow to replace pixels of low reliability.
In contrast, our evaluation on KITTI shows a larger improvement for the remaining pixels.
This correlates well with the fact that we found the output probabilities of \cite{Yin:2019:HDD} to be less well calibrated on KITTI, judging by the comparatively larger benefit of oracle confidences, \cf \cref{sec:exp_optical_flow}.
However, when comparing the relative improvements of PPACs to the ones obtained by PACs ($8.87\%$ for more reliable and $2.72\%$ for uncertain pixels), we observe that PPACs nevertheless allow for better handling of unreliable regions even if the reliability estimates are not completely accurate themselves. \begin{figure*}[t]
  \begin{subfigure}[b]{0.246\linewidth}
  \centering
  \includegraphics[width=\linewidth]{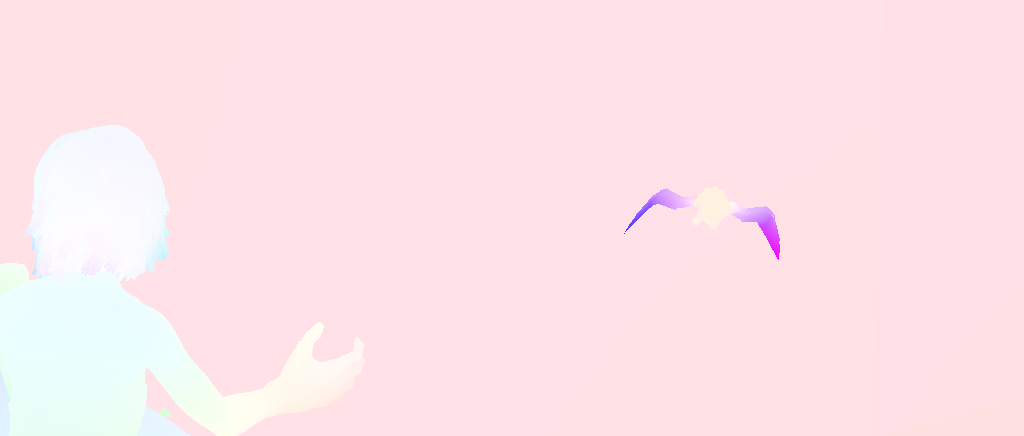}
  \end{subfigure}
  \hfill
  \begin{subfigure}[b]{0.246\linewidth}
    \centering
    \includegraphics[width=\linewidth]{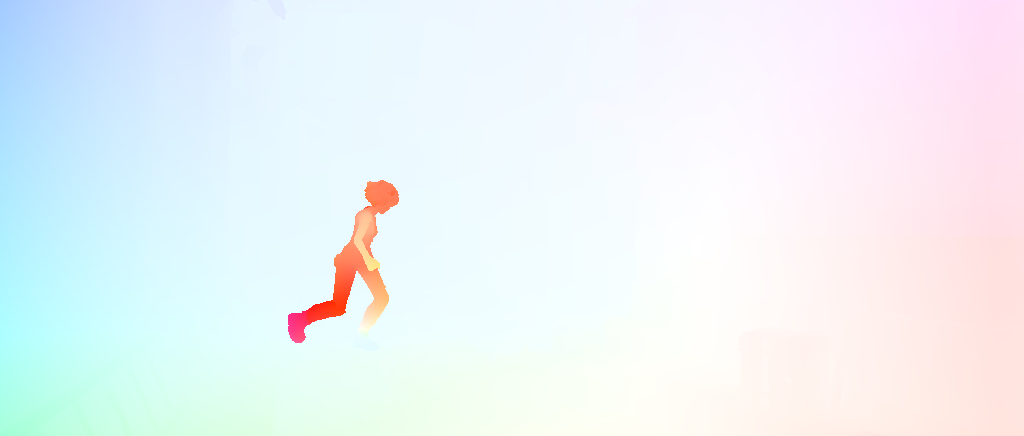}
  \end{subfigure}
  \hfill
  \begin{subfigure}[b]{0.246\linewidth}
  \centering
  \includegraphics[width=\linewidth]{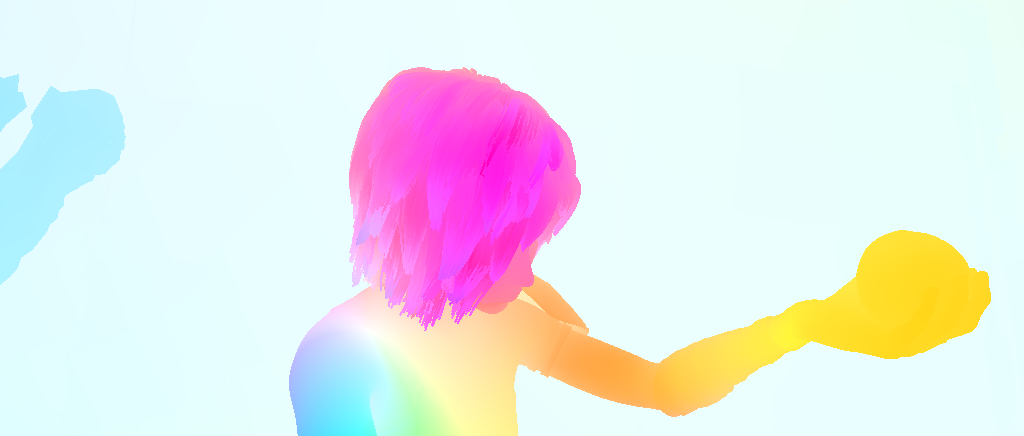}
  \end{subfigure}
  \hfill
  \begin{subfigure}[b]{0.246\linewidth}
    \centering
    \includegraphics[width=\linewidth]{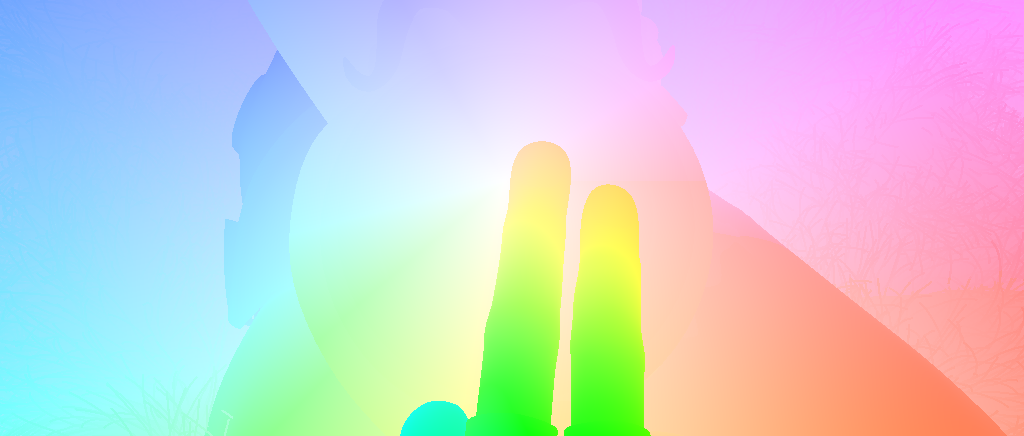}
  \end{subfigure} \\
  \begin{subfigure}[b]{0.246\linewidth}
    \centering
    \includegraphics[width=\linewidth]{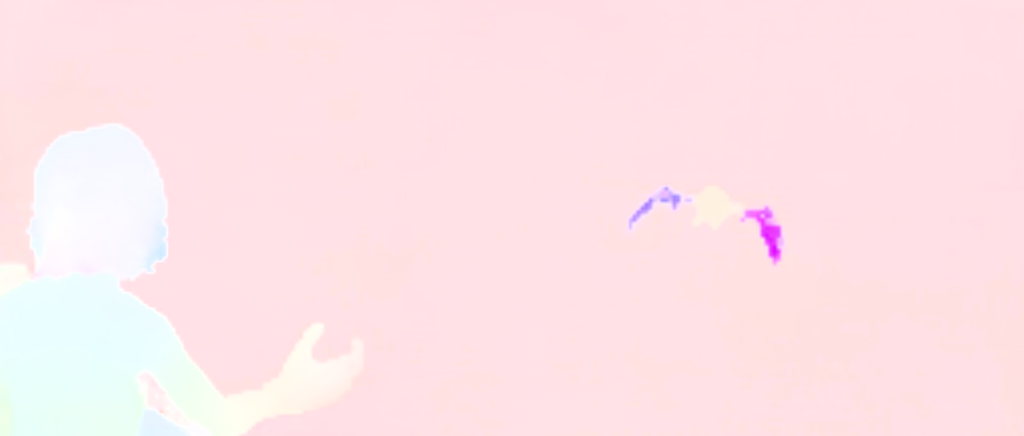}
  \end{subfigure}
  \hfill
  \begin{subfigure}[b]{0.246\linewidth}
    \centering
    \includegraphics[width=\linewidth]{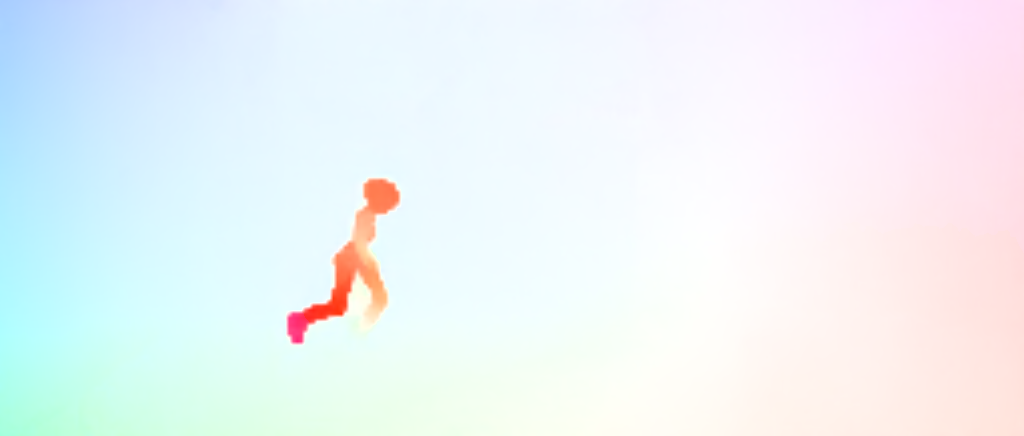}
	\end{subfigure}
  \hfill
	\begin{subfigure}[b]{0.246\linewidth}
    \centering
    \includegraphics[width=\linewidth]{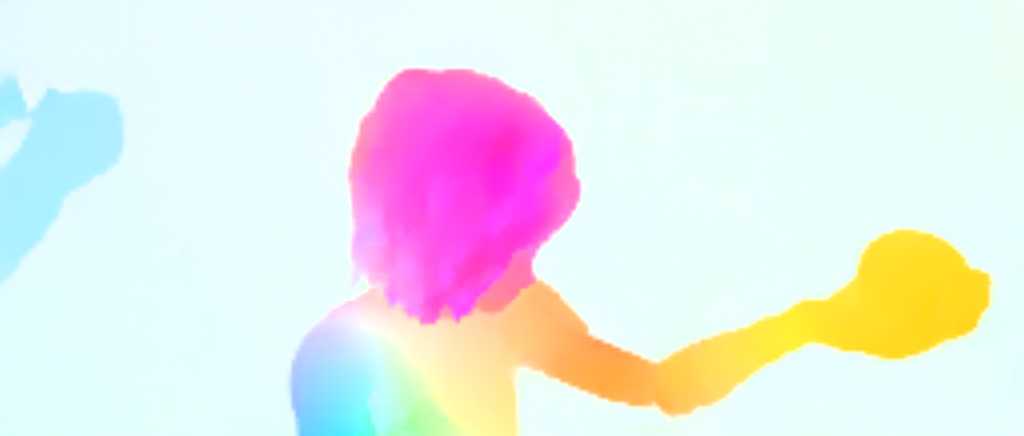}
  \end{subfigure}
  \hfill
  \begin{subfigure}[b]{0.246\linewidth}
    \centering
    \includegraphics[width=\linewidth]{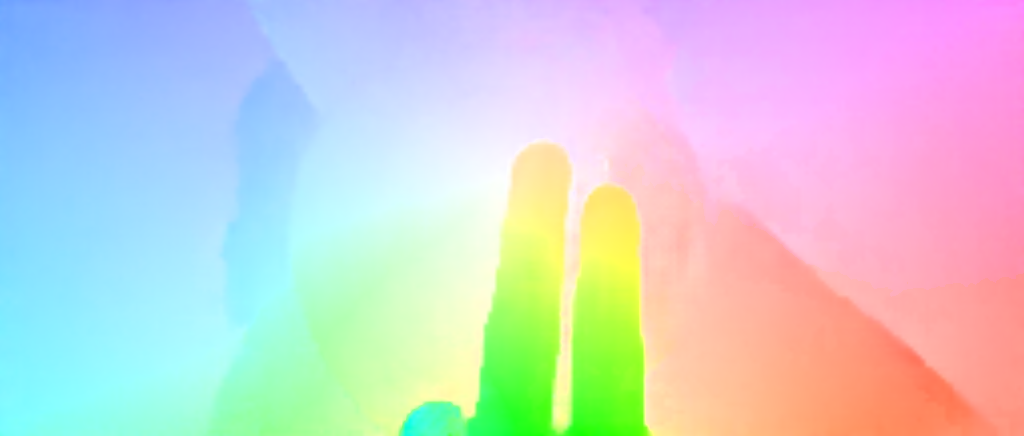}
	\end{subfigure}\\
  \begin{subfigure}[b]{0.246\linewidth}
    \centering
    \includegraphics[width=\linewidth]{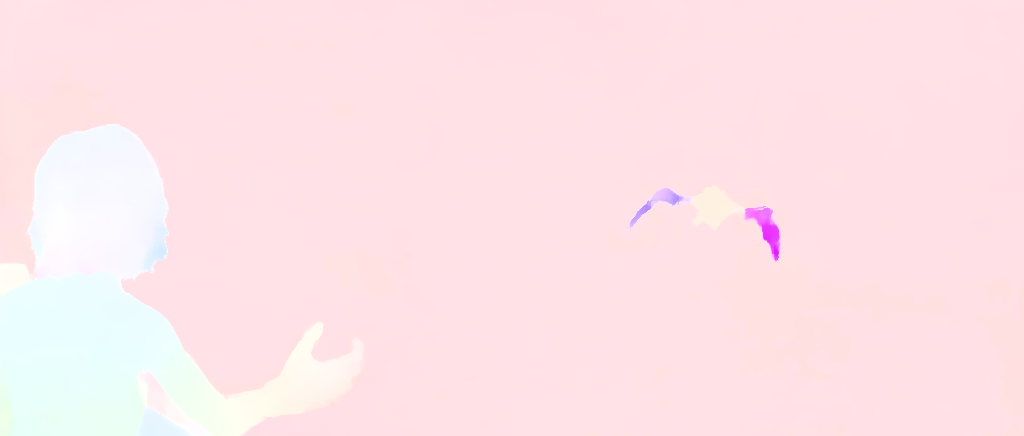}
  \end{subfigure}
  \hfill
  \begin{subfigure}[b]{0.246\linewidth}
    \centering
    \includegraphics[width=\linewidth]{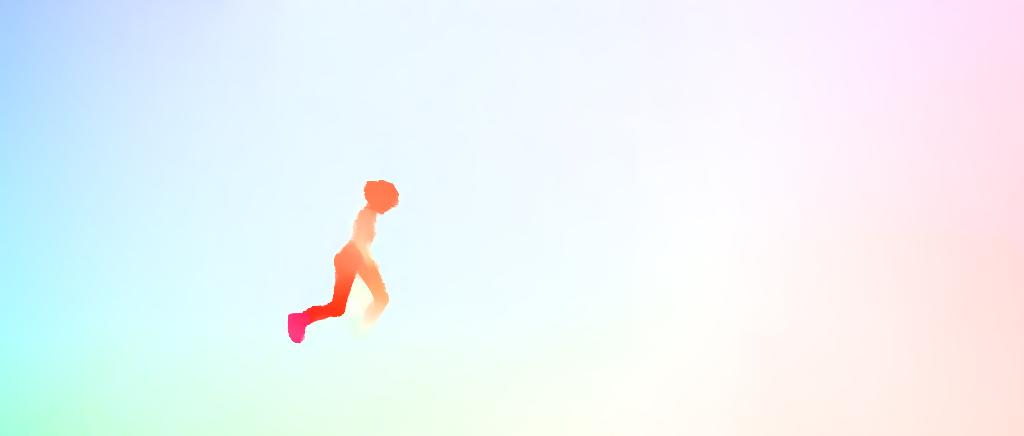}
	\end{subfigure}
  \hfill
	\begin{subfigure}[b]{0.246\linewidth}
    \centering
    \includegraphics[width=\linewidth]{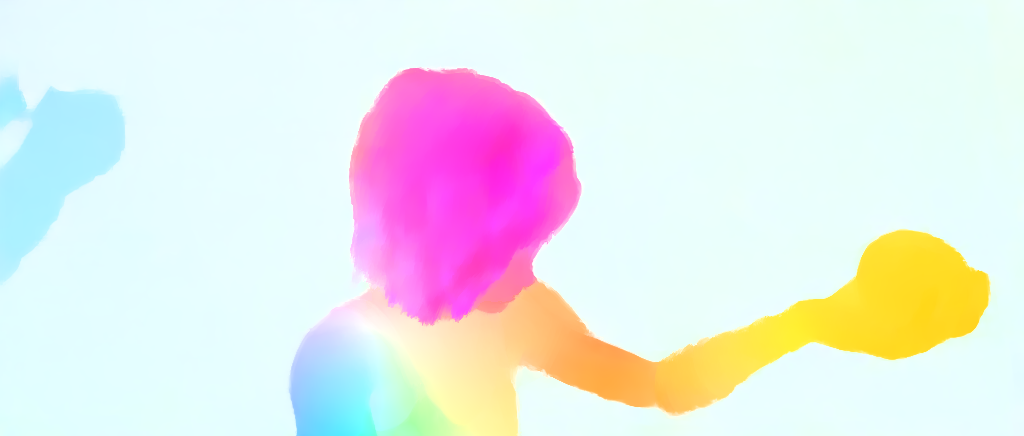}
  \end{subfigure}
  \hfill
  \begin{subfigure}[b]{0.246\linewidth}
    \centering
    \includegraphics[width=\linewidth]{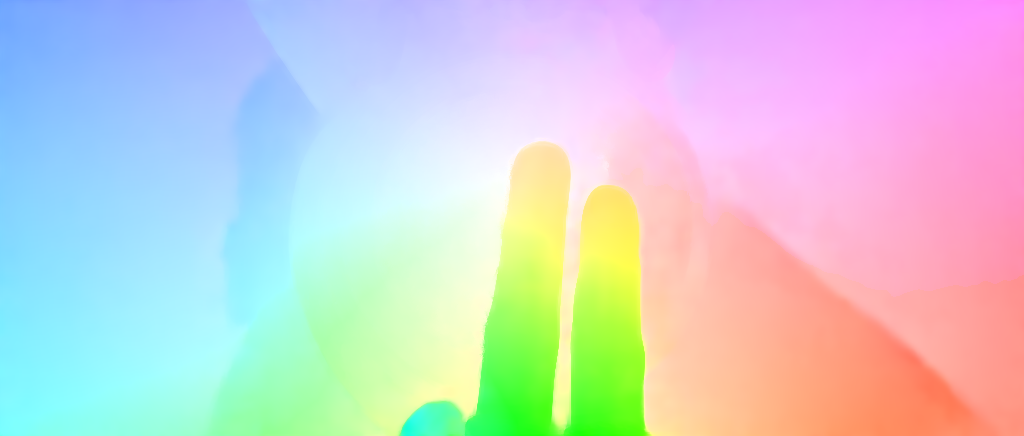}
	\end{subfigure}
  \caption{Examples of ground truth \emph{(top)}, HD3 optical flow \cite{Yin:2019:HDD} \emph{(middle)}, and our PPAC-refined optical flow \emph{(bottom)} on Sintel final. \emph{Best viewed on screen.}}
  \label{fig:examples_sintel}
  \vspace{-0.5em}
\end{figure*}
\section{Additional Visualizations}
\cref{fig:examples_sintel} shows additional visualizations of refined optical flow fields on our own validation and test splits of Sintel final.
As such, none of these flow fields was presented to the PPAC refinement network during training.
We clearly observe improved motion boundaries but also the ability of our approach to correctly propagate estimates into erroneous regions, \eg the bird wings on the leftmost example.

\begin{figure}[t]
\centering
  \begin{subfigure}[b]{0.49\linewidth}
  \centering
  \includegraphics[width=\linewidth]{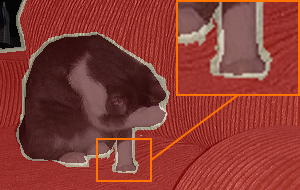}
  \end{subfigure}
  \hfill
  \begin{subfigure}[b]{0.49\linewidth}
    \centering
    \includegraphics[width=\linewidth]{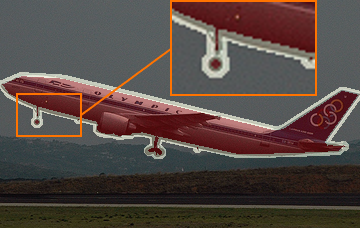}
  \end{subfigure} \\
  \vspace{0.1em}
  \begin{subfigure}[b]{0.49\linewidth}
    \centering
    \includegraphics[width=\linewidth]{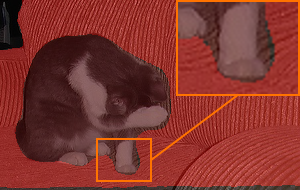}
  \end{subfigure}
  \hfill
  \begin{subfigure}[b]{0.49\linewidth}
    \centering
    \includegraphics[width=\linewidth]{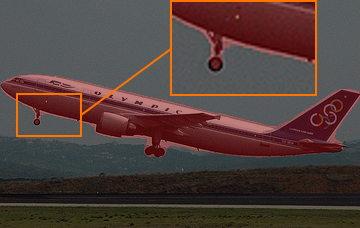}
	\end{subfigure} \\
	\vspace{0.1em}
  \begin{subfigure}[b]{0.49\linewidth}
    \centering
    \includegraphics[width=\linewidth]{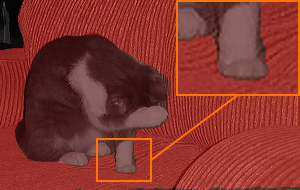}
  \end{subfigure}
  \hfill
  \begin{subfigure}[b]{0.49\linewidth}
    \centering
    \includegraphics[width=\linewidth]{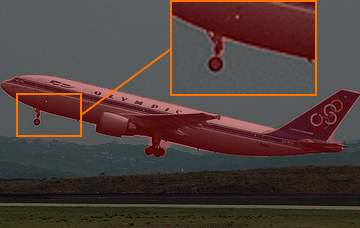}
	\end{subfigure}
  \caption{Additional examples of cropped ground truth \emph{(top)}, DeepLabv3+ \cite{Chen:2018:EAS} \emph{(middle)}, and PPAC-refined segmentation maps \emph{(bottom)} on Pascal VOC 2012. \emph{Best viewed on screen.}}
  \label{fig:additional_examples_pascal}
  \vspace{-0.5em}
\end{figure}

In \cref{fig:additional_examples_pascal}, we provide additional visualizations of refined segmentation maps on Pascal VOC 2012.
PPAC refinement leads to a clear reduction of errors near object boundaries,
\eg by considerably minimizing the segmentation margin visible at the cat paw. 
{\small
\bibliographystylesupp{ieee_fullname}
\bibliographysupp{short,litSupp}

\end{document}